\documentclass[sigconf]{acmart}

\AtBeginDocument{%
  \providecommand\BibTeX{{%
    \normalfont B\kern-0.5em{\scshape i\kern-0.25em b}\kern-0.8em\TeX}}}

\setcopyright{none}

\renewcommand\footnotetextcopyrightpermission[1]{}

\author{
Daoyuan Chen$^*$, 
Dawei Gao$^*$, 
Yuexiang Xie, 
Xuchen Pan, 
Zitao Li, \\
Yaliang Li$^\dagger$,
Bolin Ding,
Jingren Zhou \\ 
Alibaba Group
}

\usepackage{booktabs}       
\usepackage{amsfonts}       
\usepackage{nicefrac}       
\usepackage{microtype}      
\usepackage{xcolor}         
\usepackage{amsmath}
\usepackage{amsfonts}
\usepackage{amsthm}
\usepackage{graphicx}
\usepackage{color}
\usepackage{url}
\usepackage{stackrel}
\usepackage{soul}
\usepackage{multicol}
\usepackage{subfigure}
\usepackage{xspace}
\usepackage{tabularx}
\usepackage{balance}
\usepackage{wrapfig}
\usepackage{multirow}
\usepackage{booktabs}
\usepackage{tabu}
\usepackage{algorithm}
\usepackage{algpseudocode}
\usepackage{subfigure}
\usepackage{graphicx}
\usepackage{float}

\renewcommand{\algorithmicrequire}{\textbf{Input:}}  
\renewcommand{\algorithmicensure}{\textbf{Output:}} 
\usepackage{cleveref}

\newcommand{\ie}{{\em i.e.}\xspace}
\newcommand{\eg}{{\em e.g.}\xspace}

\newcommand{\squishlist}{
	\begin{list}{$\bullet$}{
		\setlength{\itemsep}{0pt}
		\setlength{\parsep}{3pt}
		\setlength{\topsep}{3pt}
		\setlength{\partopsep}{0pt}
		\setlength{\leftmargin}{1.0em}
		\setlength{\labelwidth}{1em}
		\setlength{\labelsep}{0.5em}
   }
}

\newcommand{\squishenum}{
	
	\begin{list}{\usecounter{scount}}{
		\setlength{\itemsep}{0pt}
		\setlength{\parsep}{3pt}
		\setlength{\topsep}{3pt}
		\setlength{\partopsep}{0pt}
		\setlength{\leftmargin}{1.2em}
		\setlength{\labelwidth}{1em}
		\setlength{\labelsep}{0.5em}
	}
}

\newcommand{\squishend}{
	\end{list}
}

\newcommand{\oursys}{\textsc{FS-Real}\xspace}

\usepackage[textsize=tiny]{todonotes}

\usepackage{amsmath,amsfonts,bm}

\makeatletter
\def\widebreve{\mathpalette\wide@breve}
\def\wide@breve#1#2{\sbox\z@{$#1#2$}%
	\mathop{\vbox{\m@th\ialign{##\crcr
				\kern0.08em\brevefill#1{0.6\wd\z@}\crcr\noalign{\nointerlineskip}%
				$\hss#1#2\hss$\crcr}}}\limits}
\def\brevefill#1#2{$\m@th\sbox\tw@{$#1($}%
	\hss\resizebox{#2}{\wd\tw@}{\rotatebox[origin=c]{90}{\upshape(}}\hss$}
\makeatletter

\def\eqref#1{equation~\ref{#1}}

\def\1{\bm{1}}

\DeclareMathAlphabet{\mathsfit}{\encodingdefault}{\sfdefault}{m}{sl}
\SetMathAlphabet{\mathsfit}{bold}{\encodingdefault}{\sfdefault}{bx}{n}

\begin{document}

\title{\oursys: Towards Real-World Cross-Device Federated Learning}

\begin{abstract}
Federated Learning (FL) aims to train high-quality models in collaboration with distributed clients while not uploading their local data, which attracts increasing attention in both academia and industry.
However, there is still a considerable gap between the flourishing FL research and real-world scenarios, mainly caused by the characteristics of heterogeneous devices and its scales.
Most existing works conduct evaluations with homogeneous devices, which are mismatched with the diversity and variability of heterogeneous devices in real-world scenarios.
Moreover, it is challenging to conduct research and development at scale with heterogeneous devices due to limited resources and complex software stacks.
These two key factors are important yet underexplored in FL research as they directly impact the FL training dynamics and final performance, making the effectiveness and usability of FL algorithms unclear.
To bridge the gap, in this paper, we propose an efficient and scalable prototyping system for real-world cross-device FL, \oursys. 
It supports heterogeneous device runtime, contains parallelism and robustness enhanced FL server, and provides implementations and extensibility for advanced FL utility features such as personalization, communication compression and asynchronous aggregation. 
To demonstrate the usability and efficiency of \oursys, we conduct extensive experiments with various device distributions, quantify and analyze the effect of the heterogeneous device and various scales, and further provide insights and open discussions about real-world FL scenarios. 
Our system is released to help to pave the way for further real-world FL research and broad applications involving diverse devices and scales.

\end{abstract}
\maketitle
\pagestyle{plain}

\renewcommand*{\thefootnote}{\fnsymbol{footnote}}
\footnotetext[1]{Co-first authors.} 
\footnotetext[2]{Corresponding author, email address: yaliang.li@alibaba-inc.com.}
\renewcommand*{\thefootnote}{\arabic{footnote}}
\renewcommand{\shortauthors}{Daoyuan Chen, Dawei Gao, Yuexiang Xie, Xuchen Pan, Zitao Li, Yaliang Li, Bolin Ding, and Jingren Zhou}
\renewcommand{\shorttitle}{\oursys: Towards Real-World Cross-Device Federated Learning}

\section{Introduction}
\label{sec:intro}

Cross-device federated learning (FL) aims to leverage a large-scale distributed group of clients to collaboratively train high-quality machine learning models, while retaining the client data locally to devices for privacy protection \cite{fedavg, yang2019federated, advancesopenfl,konevcny2016federated, nguyen2021federated}. 
In recent years, the boom of edge intelligence and the growing demand for data privacy protection has spawned remarkable innovations in FL algorithms \cite{lim2020federated, li2020review, mothukuri2021survey}. 
However, there is still a considerable gap between cross-device FL research and practical solutions in real-world application scenarios, particularly in terms of runtime characteristics and the scale of the participated devices.
\textit{Without bridging the gap soon, the effectiveness and usability of both existing and follow-up FL research are challenged.}

To be specific, let us start by explaining why the gap exists in FL:
\textbf{(a) Homogeneous v.s. Heterogeneous Device}. Most existing studies conduct FL experiments based on homogeneous device settings: each participated device has the same computational, storage, and communication capabilities \cite{advancesopenfl, li2020federated}. 
In real-world applications, however, the device capabilities are highly heterogeneous and dynamically changing. For example, the computational power of mainstream smartphones can vary by orders of magnitude \cite{mobile-bench,wu2019machine}. 
Further, the available computational and communication resources of an FL device can change due to the competition from other apps on the same device and the spotty network connection \cite{flsurveyiot}. 
\textbf{(b) Device Scales.} Furthermore, it is highly challenging to conduct research at scale with real heterogeneous devices. Scaling heterogeneous devices requires researchers to deal with diverse device hardware and software environments to reflect real computing capacities, and build distributed communications through the network interface cards to reflect real transmission capabilities \cite{flsyssurvey}. 
Due to limited resources and complex software stacks, most existing works focus on standalone simulation studies with high-performance servers, which introduces non-negligible simulation errors compared with real scenarios. For more details, please see the observations in Section \ref{exp:oursys}.

Notably, the impact of these two key factors on recent advanced FL algorithms is non-trivial while still under-explored. 
For example, in practical FL scenarios, it is important to get a high-performance model in a short time and with few resources.
Personalized FL shows promising results to improve the performance when dealing with Non-IID data \cite{towardspfl, pflbench, zhu2021federated};
communication compression \cite{Haddadpour2020FederatedLW, konevcny2016federated2,wang2022communication} and asynchronous aggregation \cite{nguyen2022fedbuff, xie2019asynchronous, chen2020asynchronous} can effectively accelerate FL training speed by reducing traffic and increasing device utilization respectively.
However, \textit{there are inconsistent training dynamics, convergence time and trained model of FL with heterogeneous and homogeneous devices}, which are closely related to the performance of individual clients.
Furthermore, such inconsistency is exacerbated by various real-world scales of FL devices, calling for further validation of effectiveness and usability of existing solutions.

To bridge this gap, in this paper, we propose a prototyping system for real-world FL, \oursys, based on which we aim to identify major challenges in real heterogeneous device and scalable FL scenarios, providing reusable functionality and valuable insights for further FL research, development and deployment.
Specifically, \oursys contains a flexible heterogeneous device runtime, a group of efficient and scalable FL device executors and FL server, and implementations for a number of practical and advanced FL enhancement techniques with easy extensions.
(1) The \oursys runtime enables users to easily and cost-efficiently study FL performance in real FL scenarios, where the devices can have diverse scales and highly different hardware capabilities, e.g., with configurable computation and communication resources such as CPU cores and network types.
(2) Based on a computation engine optimized for edge intelligence, MNN \cite{proc:osdi22:walle}, we implement efficient lightweight executors for FL behaviors such as local training, which can be executed in real Android phones and IoT devices. 
To enhance the training efficiency and scalability, we carefully decouple and schedule the FL plan so that server-side processing behaviors such as sending, receiving, and aggregation of information can be efficiently parallelized and scaled up. 
(3) Further, we incorporate several representative personalized FL, communication compression, and asynchronous FL algorithms into \oursys, and provide simple and easily customizable programming interfaces for future extensions.

With the implemented \oursys, we conduct an extensive evaluation to demonstrate its usability, efficiency, and scalability. We first examine the performance of existing FL algorithms with different hardware device distributions and varied device scales.
We find that both distribution differences and quantity differences in heterogeneous devices bring a substantial model performance gap between homogeneous and heterogeneous scenarios, and such gap becomes more noticeable in terms of fairness-related metrics and at large device scales. These findings confirm the strong need of \oursys.
Besides, heterogeneous devices exhibit complex differences in convergence speed, network traffic and client utilization, which are specific to the device scales and device distributions, challenging the utility of FL algorithms in real-world cross-device federated learning scenarios. 
Moreover, we find advanced FL techniques, such as personalized FL, communication compression, and asynchronous aggregation, work well in most evaluated heterogeneous device cases, while the performance gain of these techniques suffers from high variance especially at large scales, calling for future attention to their usability and robustness.

Our contributions are summarized as follows: 
\begin{itemize}
	\item We propose an efficient and scalable prototype system \oursys for real-world cross-device FL, which supports heterogeneous device runtime and includes several advanced FL utility features such as personalization, communication compression, and asynchronous concurrency. 
	\item With experiments conducted on up to thousand-scale heterogeneous devices (android phones), we quantify and analyze how the underexplored factor, heterogeneous device, affects the FL performance under different scenarios and different device distributions, and point out some challenges and open issues when scaling up FL.
	\item We release the system at https://github.com/alibaba/Federated
 Scope/tree/FSreal. With the open-sourced system and provided insights, we hope that our work can greatly facilitate further research and broad applications on real-world FL scenarios that would otherwise be infeasible without a dedicated real system.
 	
\end{itemize}

\section{Background and Related Work}
\label{sec:backgrond-related}

\subsection{FL Algorithms}
A great deal of effort has been devoted to FL algorithm research \cite{wang2021field,li2020federated, advancesopenfl, nguyen2021federated}. 
Typically, FL algorithms adopt the following workflow similar to the well-known FedAvg \cite{fedavg}: 
At the $t$-th FL round, the server first selects $n$ available clients $\mathcal{C}_{ava}$ from all the $N$ participated clients, and sends the current global model weight $\bm{\theta_g^{t}}$  to the selected clients.
Then the selected client $i$ trains the received model with their local private data, and uploads the updated weight $\bm{\theta_i^{t+1}}$ to the server.
Finally, the server aggregates the model updates from $n'$ responded clients $\mathcal{C}_{res} \subseteq \mathcal{C}_{ava}$ to generate the next-round global model as
$\bm{\theta_g^{t+1}}= \sum_{i=1}^{n'}w_i\bm{\theta_i^{t+1}}$ where $w_i$ is the aggregation weight that is usually defined as the ratio of the number of training data of client $i$ and the total number for all the $n'$ clients.
This process is repeated until $\bm{\theta_g}$ converges or the round reaches the pre-defined maximum number $T$.

In real-world FL scenarios, $N$ can be in various scales and even be very large with highly heterogeneous data distribution, and the clients usually have limited device resources and spotty connections.
These pose difficulties to learn high-quality FL models with fast convergence speed and low resource cost. 
Researchers have proposed fruitful algorithms to improve FedAvg such as personalization with client-specific model $\bm{\theta_i}$ \cite{towardspfl,fedbn,li2021ditto,oh2022fedbabu, fedrep}, heterogeneity-aware sampling protocols \cite{fedbalancer,Oort-osdi21}, communication compression \cite{konevcny2016federated, Haddadpour2020FederatedLW, xu2020ternary} and asynchronous aggregation \cite{chen2020asynchronous, xie2019asynchronous, lu2019differentially}.
However, most existing FL algorithms are studied and validated in homogeneous device simulation environments, in which the responded clients $\mathcal{C}_{res}$ at each round will \textit{differ} from the one in heterogeneous device setting, and subsequently leads to \textit{inconsistent} available clients $\mathcal{C}_{ava}$,  convergence round $T'$ and aggregated model sequence $\{\bm{\theta_g^2, \theta_g^3, \cdots, \theta_g^{T'}}\}$. 
In this work, different from most existing homogeneous device works, we focus on real heterogeneous device scenarios and study the effect of various heterogeneous device scales.

\subsection{FL Systems and Tools}
There are a number of FL systems and tools that make efforts to translate academic FL progress into real-world or scalable solutions.
Some industrial systems have performed FL on real heterogeneous runtime mobile devices with $n'$ in the hundreds to thousands scale, such as Google FL stack \cite{googlefl}, Papaya proposed by Facebook \cite{papaya}, and Apple FL \cite{afl}.
However, these works are closed-source.
Among the many open-source frameworks, benchmarks and tools such as LEAF \cite{leaf}, TensorFlow Federated \cite{tff}, PySyft \cite{pysyft}, FedML \cite{fedml}, Flower \cite{flower}, OpenFL \cite{openfl_citation}, IBM FL \cite{ibmfl},  FederatedScope \cite{fs}, FLARE \cite{nvidiaFlare}, large-scale realistic heterogeneous device runtime is not yet well supported.
Ideally, one can approximate the response time of an individual device as $|\bm{\theta}|/B_{up} + |\bm{\theta}|/B_{down} + |D|s + t_{other}$, where $|\bm{\theta}|$ indicates the model size, $B_{up}$ and $B_{down}$ indicates the upload and download speeds respectively, $|D|$ is the size of the local training data, $s$ is the training speed and $t_{other}$ indicates the total delay of other processings such as I/O. 
Any one of these factors can vary by orders of magnitude on different devices (e.g. $B_{up}$ on a 4G network can be 200 times faster than on a 3G network) and consequently cascade to cause huge inter-device response differences. In addition, some device states tend to change dynamically and become unavailable according to conditions defined by specific FL protocols and applications, such as network type switching, FL executor being shut down by the user or OS, device running low on power, etc.
FedScale \cite{fedscale} and FLASH \cite{flash} consider introducing heterogeneous message arrival times with cost models and virtual timestamps, but their fidelity is still limited by the capability of the cost models, and the precision and coverage of the device capacity values (\eg, $s$ and $t_{other}$ mentioned above).
Our work differs from theirs by supporting real and scalable heterogeneous device runtime with an efficient device training engine, and high-fidelity simulation with different mobile types and device distributions.

Besides, these above works mainly focus on functionality and usability of system or tool.
Different from them, we studied the impact of different device runtime distributions and scales on the FL, especially in conjunction with some of the recent FL algorithms such as personalization, compression, and asynchronous aggregation.

\section{The Proposed \oursys}
In this section, we describe the design and implementation of our system optimized for real-world FL.

\subsection{Design Principles}
\label{sec:design-require}
The two main characteristics of real cross-device FL (\ie, \textit{heterogeneous device} and \textit{large scale}) impose a number of unique system requirements in terms of the following aspects:
\textbf{(i) Usability and Efficiency.} The hardware and software environments of devices are very different from cloud-based high-performance servers, having various instruction sets (x86, ARM, etc.,), operating systems, and library dependencies. 
A cross-device FL system should have good usability and can conduct evaluation for a wide range of device runtimes.
Besides, devices often have limited resources, such as computing capacity, communication bandwidth, and storage.
It is critical for cross-device FL systems to efficiently conduct training, inference, and management for local models with minimal consumption of device resources.
\textbf{(ii) Scalability and Robustness.} As the participating devices can be on diverse scales, the FL server should be highly scalable, make good use of system resources, and yield a corresponding improvement in model quality and training speed as more system resources are invested. 
Moreover, in real scalable FL scenarios with heterogeneous devices, many devices are prone to be slow and disconnected as we discussed aforementioned.
How to robustly handle such devices is also one of the key challenges.
\textbf{(iii) Flexibility and Extensibility.} 
Real FL applications require the collaboration between devices and the server, which involves a large number of potentially inconsistent software stacks and programming interfaces.
Supporting flexible customization and extension of FL algorithms is therefore necessary to improve the model quality and convergence for diverse scales and scenarios.

Next we introduce how we implement the components to fulfill the above requirements (\textbf{RS}), including an easy-to-use and high-fidelity simulation platform and efficient FL device executors (Sec.\ref{sec:heter-runtime} for \textbf{RS i}); a scalability enhanced robust server (Sec.\ref{sec:server} for \textbf{RS ii}); and supports and extensions of diverse advanced FL algorithms (Sec.\ref{sec:fl-algo} for \textbf{RS iii}).

\subsection{Heterogeneous Device Runtime}
\label{sec:heter-runtime}
For the sake of usability and efficiency, we implement a dedicated high-performance FL execution engine for heterogeneous devices. 
The major modules are designed as a portable learner based on the MNN \cite{proc:osdi22:walle}, and a communication and storage manager based on the native APIs of the target OS (e.g. Android SDK).
The learner is responsible for local model training and monitoring, and can be selectively packaged and compiled into lightweight dynamic link libraries for different target hardware (e.g. x86 CPUs, ARM CPUs, and CUDA-GPU).
We leverage MNN to automatically optimize the computational graphs for the target hardware, such that the computation and memory access can be accelerated, and the binary size can be reduced. 

Besides, the communication and storage manager is responsible for sending messages to (and receiving messages from) the server and for the serialization of the data model, in which we leverage gRPC and compressed MNN models to enable high-performance network transmission.
Here we use the MNN model file as an intermediate representation of the exchanged model, which facilitates cross-platform execution and reduces development costs. With support for mainstream model formats such as onnx \cite{onnx}, Tensorflow \cite{tensorflow}, Torchscripts \cite{pytorch} and a large number of widely used operations, we can easily define computational graphs on the server side using python and various frameworks, which will be automatically and uniformly converted to MNN model on devices, reducing the programming burden for diverse devices.

Our efficiency optimization is multi-granular in terms of (1) the device learning behaviors with efficient C++ implementation and leveraging of heterogeneous computing hardware (e.g., high-performance assembly codes for different operators with the help of MNN), (2) the communication and storage manager implemented with native OS API to reduce redundant memory accesses and computational calls from other intermediate code bases and (3) the compact binary size and minimal dependency that reduce the burden on users' storage resources. In Sec.\ref{exp:oursys}, we will give some quantitative comparisons to demonstrate the efficiency of \oursys.

\subsection{Enhanced FL Server}
In real-world FL scenarios, the FL server can easily become a performance bottleneck for the whole system due to the fact that as the scale of clients increases, both the consumed resources and the number of slow or disconnected devices increase. 
In this section, we introduce the enhanced FL server in \oursys for better scalability and robustness.
\label{sec:server}
\subsubsection{Message Concurrency.}
To enhance the scalability of the system, we implement concurrency at multiple granularities. 
We choose a message-passing-based FL library FederatedScope \cite{fs} as a starting point for the server implementation, and further extend and optimize it.
By focusing on the unified message object, we can easily analyze and handle potential space-time contention and redundant resource overheads.
Specifically, we first abstract FL server behaviors into \textit{Message Transmitter} that is responsible for the transfer of messages between clients and server, and \textit{Message Processor} that is responsible for a series of FL transactions to process messages, such as model aggregation and monitoring.
We implement \textit{Message Transmitters} and \textit{Message Processors} with multiple individual processes for interleaved execution, due to the fact that they typically occupy different hardware resources (e.g., CPU, Network Interface Card and disks) and have great potential for concurrency.
Note that this design and optimization is necessary and effective because the transmission latency and processing latency of messages can vary by even several orders of magnitude in different FL and network situations.
Furthermore, for transmitters, we introduce separate process pools of configurable size for parallel gRPC sending and receiving, so that they can be flexibly and automatically scale out to efficiently handle different scales of FL participated devices.

\subsubsection{Robust Client Selection.}
The FL server needs to select the available devices and broadcast messages to them to initiate local training in each round, perform aggregation at the proper time according to the response messages, and subsequently start the next round of broadcasting. In real FL scenarios, the response times between devices can differ by order of magnitude as aforementioned.

To advance the FL process robustly and efficiently in various scenarios, we design a timeout mechanism for FL workload adaptation. 
Suppose that a server in an FL round has selected $n$ devices from $n_{ava}$ available devices to broadcast messages and expects to receive $n'$ device responses for aggregation.
Given a timeout budget $t_o$, if the number of received messages is less than $n'$ within $t_o$ time after broadcasting, we will rebroadcast the messages to other $min(n, max(n_{ava}-n, 0))$ available devices and reset the timeout timer.
Inspired by the AIMD (Additive Increase, Multiplicative Decrease) congestion control algorithm \cite{tcpaimd}, we double $t_o$ after each triggered rebroadcast and subtract $\delta_t$ from $t_o$ when no rebroadcast is triggered for $k$ consecutive rounds.
This mechanism takes into account the diverse load and dynamics of the available heterogeneous devices, allowing for efficient advancement of FL process across time and scenarios based on the overall responsiveness of the devices.
Besides, we support the over-selection mechanism on the server side, by broadcasting to $min(n_{ava}, \lfloor qn \rfloor)$ available devices, where $q$ indicates the over-selection rate (e.g. 150\%). This mechanism enhances the robustness of synchronous aggregation scenarios, while in the next section, we describe asynchronous algorithms that can make FL algorithms more efficient and robust.

\subsection{Advanced FL Techniques}
\label{sec:fl-algo}
In order to provide flexible and easy FL algorithm extensions, we design simple and unified configuration options and programming interfaces for both the cloud and the device side in \oursys.
These APIs abstract a set of expressive behaviors for common yet necessary objects in FL such as messages, models, trainers, monitors, etc., which are decoupled from the target device runtime and can be flexibly customized.
In particular, here we support and provide a number of advanced FL technology implementations that are important for scaling heterogeneous devices FL, including personalization, communication compression, and asynchronous aggregation.

(1) Personalization is important to improve the quality of client-specific models and ensure the user experience of end-intelligence applications where each device corresponds to a single real user. 
In the \textit{Device Executor} (Sec.\ref{sec:heter-runtime}), we provide device-wise configuration and interfaces such as ``set\_local\_module'', ``fine\_tune'' and implement SOTA personalized FL algorithms such as personalized fine-tuning, FedRep \cite{fedrep} and FedBABU \cite{oh2022fedbabu}.
(2) Communication compression reduces the message size to save the running time and reduce development costs (such as the fee charged for mobile traffic).
In the server's \textit{Message Transmitter} (Sec.\ref{sec:server}), we provide interfaces such as ``channel\_compress'', ``para\_compress'' and support compression techniques such as lossless gzip and deflate algorithm \cite{deutsch1996gzip}, and lossy half-precision floating point (FP16) and INT2$\sim$INT8 quantization \cite{liang2021pruning}.
(3) Asynchronous aggregations combine messages received in different FL rounds to improve training efficiency and robustness. In the server's \textit{Message Processor}, we provide convenient maintenance of message staleness and necessary interfaces such as ``stale\_aggregation'' for asynchronous algorithms, and implement the SOTA algorithm, FedBuff \cite{nguyen2022fedbuff}.

\subsection{Heterogeneous Device Simulation Platform}
\label{sec:heter-simula}
Preparing the corresponding hardwares (\eg, mobile phones) for heterogeneous runtimes to run real FL requires high costs in terms of development time, financial expense for acquiring new hardware devices and potential loss of user experience (\eg, testing with users' devices).
To accelerate real-world FL research and reduce the incurred costs, we further implement a high-fidelity and easy-to-use heterogeneous runtime simulation platform in \oursys.

As mentioned in Sec.\ref{sec:backgrond-related}, the computing, communication, and storage capabilities of the participated devices have the most direct impact on FL, thus we mainly focus on fidelity to these aspects. 
We run separate processes for the participated clients based on the Android official native simulation tool \footnote{https://developer.android.com/studio/run/emulator}, and allow users to configure the number of CPU cores, memory footprint, network types, and network latency. 
Note that with our simulation platform, each device runs in the same software environment (\eg, OS and library dependencies) as a real phone and will communicate realistically across machines via gRPC, which greatly reduces the gap between emulation and real FL.

In addition, we provide easy-to-use configurable tools to quickly start simulated heterogeneous devices and run FL at scale with a single command. To enhance the utilization of simulation host resources, we introduce a scalable, self-managing device pool that supports different types of heterogeneous runtime distributions and allocates available devices to participated clients at each FL round. 
This configuration system is comprehensive: 
At the global level, users can specify common FL settings (e.g. dataset, FL scale, network topology, FL algorithm); 
At the local client level, different personalized hyperparameters and data configurations can be specified for each client; 
At the device level, different device capabilities (e.g. CPU, network type) can be specified for each emulated device, as well as the device pool distributions (we will give some examples in Sec.\ref{sec:exp-setting}). 
Hyperparameter optimization is also supported with early stopping and easy configuration.

Based on the available host resources and specified configurations such as the number of devices and device capacity, the device pool will automatically perform auxiliary engineering actions such as Android phone startup, \oursys runtime (App) installation, port mapping, network connecting, clients' data switching and storage, failure reboot, etc., allowing users to only focus on emulation configuration and not worry about the complexity of dealing with heterogeneous devices.
Here the device allocation can be randomized to simulate the changes in device runtime (e.g., CPU contention and network bandwidth fluctuations).

\begin{table*}[h!]
\centering
\caption{Comparisons of on-device FL training time (seconds). 
We train a ConvNet2 model on FEMNIST for 50 training rounds and use the same local data size each round.
The marker ``$\star$'' indicates the estimated value of FedScale according to its original cost model $(\#Sample \cdot LatencyPerSample) + ModelSize/Bandwidth$, which is a fixed value for different configurations.}
\label{tab:device-time}
\begin{tabular}{cccccccc} 
\toprule
\multirow{2}{*}{CPU Loads} & \multirow{2}{*}{Wifi Bandwidth} & \multirow{2}{*}{Batch Size} & \multicolumn{4}{c}{Running Time per FL Round (Seconds)} \\ 
\cmidrule(lr){4-7}
\multicolumn{1}{c}{} & \multicolumn{1}{c}{} & \multicolumn{1}{c}{} & \multicolumn{1}{c}{\oursys, Real} & \multicolumn{1}{c}{\oursys, Simula.} & FedScale, Real & FedScale, Simula. \\ 
\midrule
\multirow{3}{*}{Idle} & \multirow{3}{*}{100Mbps} & 16 & 3.58 \textasciitilde{} 3.77 & 1.99 \textasciitilde{} 3.12 & 7.24 \textasciitilde{} 7.75 & \multirow{7}{*}{\begin{tabular}[c]{@{}c@{}}\\ 5.85$^\star$ \end{tabular}} \\
 &  & 32 & 3.42 \textasciitilde{} 3.62 & 1.70 \textasciitilde{} 2.45  & 6.33 \textasciitilde{} 6.94 &  \\
 &  & 64 & 3.31 \textasciitilde{} 3.51 & 1.54 \textasciitilde{} 2.10  & 5.84 \textasciitilde{} 6.63 &  \\ 
\cmidrule(lr){1-6}
Idle & 25Mbps & 32 & 8.66 \textasciitilde{} 9.24 & 3.31 \textasciitilde{} 4.54  & 12.28 \textasciitilde{} 13.27 &  \\
Idle & 5Mbps & 32 & 43.7 \textasciitilde{} 53.1 & 20.31 \textasciitilde{} 28.98  & 51.55 \textasciitilde{} 50.18 &  \\ 
\cmidrule(lr){1-6}
Light & 100Mbps & 32 & 3.48 \textasciitilde{} 3.54 & 3.07 \textasciitilde{} 5.76 & 6.53 \textasciitilde{} 7.44 &  \\
Moderate~ & 100Mbps & 32 & 3.66 \textasciitilde{} 4.05 & 2.12 \textasciitilde{} 3.56 & 7.84 \textasciitilde{} 8.94 &  \\
\bottomrule
\end{tabular}
\end{table*}

\section{How \oursys benefits real FL?}
\label{exp:oursys}
In this section, we conduct experiments to show that the proposed \oursys is usable, efficient and scalable to cross-device FL for both research and deployment.

\begin{figure}[h!]
	\centering
	\includegraphics[width=0.47\textwidth]{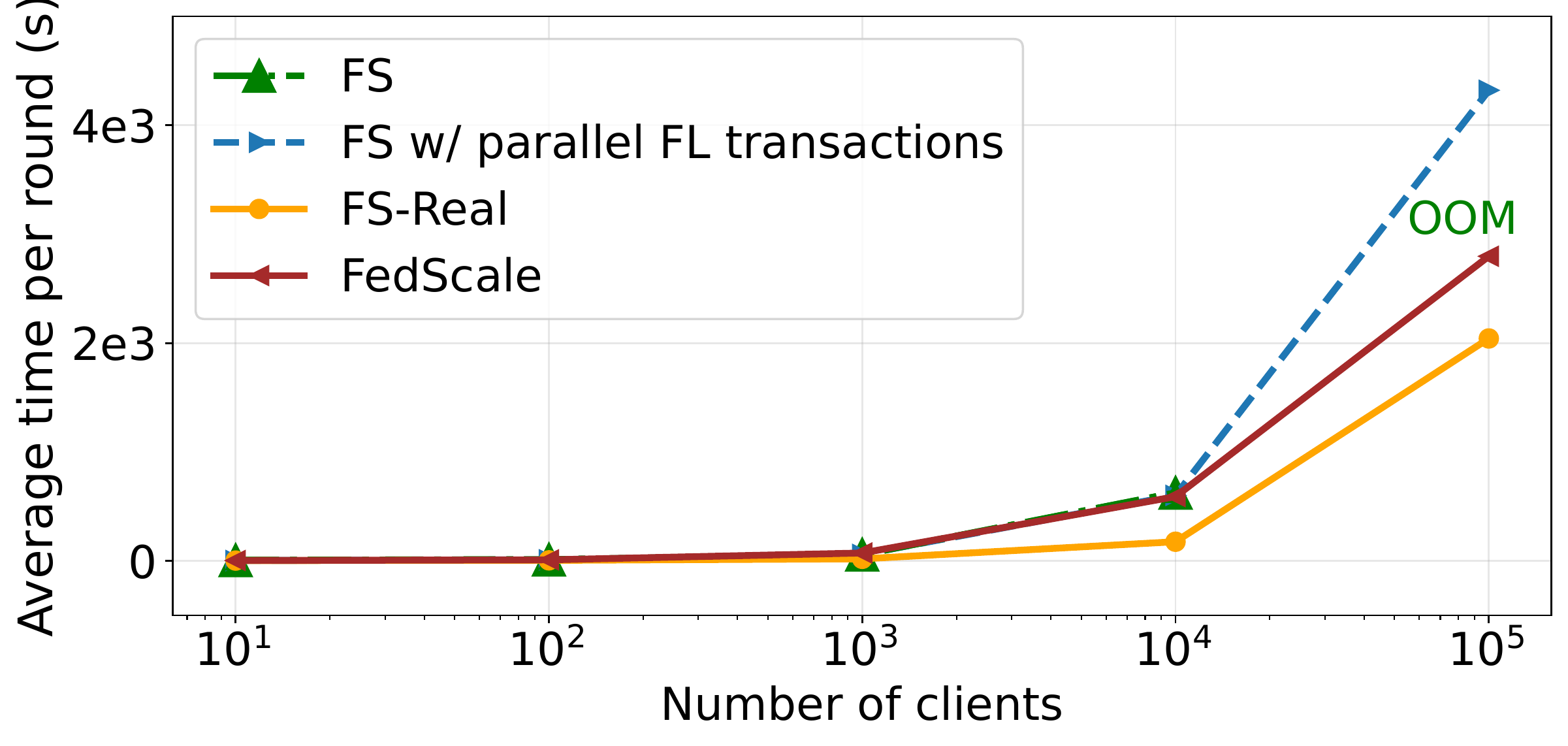}
	\caption{The scalability study of \oursys.} 
	\label{fig:our-scale}
\end{figure}

\subsection{Heterogeneous Device Runtime}
To demonstrate the usability and efficiency of the proposed \oursys for hetero-device runtime (introduced in Section \ref{sec:heter-runtime}), we conduct a series of experiments with \oursys on both real mobile phones (RedMi K40) and simulated Android devices, denoted as ``\oursys, Real'' and ``\oursys, Simula.'' respectively.
We vary the configurations of CPU loads (idle/light/moderate), network bandwidth (100Mbps/25Mbps/5Mbps), and batch sizes (16/32/64) to mimic diverse FL environments. 
For real mobile phones, a light CPU load indicates that a video player and a music player are running, and a moderate CPU load indicates that a large game is running and high-quality lighting effects are turning on. 
For simulated Android devices, we adjust the available CPU cores to imitate different CPU loads (4/2/1 CPU cores for idle/light/moderate CPU loads).
Meanwhile, we compare a representative FL tool, FedScale \cite{fedscale}, which also supports FL deployment (denoted as ``FedScale, Real'') and simulation (denoted as ``FedScale, Simula.'') for heterogeneous devices.

The results of the running time per FL training round are reported in Table \ref{tab:device-time}. 
Based on the results, we can conclude that \oursys outperforms FedScale in terms of both the \emph{efficiency} of real-world application and the \emph{fidelity} of simulation.
The reason is that \oursys allows users in running FL on real mobile devices based on MNN (with C++ codes), while FedScale is based on the Termux App to run Linux on Android (with Python codes).
We also note that \oursys requires only a portable app with 163 MB to run FL on the device, while the FedScale runtime takes up to 8.98 GB overhead, including a number of dependencies such as installed Linux OS, Numpy, and Pytorch.
(2) For fidelity of the simulation, we can observe that the simulation tool provided in \oursys can faithfully reflect the real-world diverse, changeable device runtimes within \textit{ranges}, under different computing and communication configurations (CPU cores, network bandwidths, and bath sizes).
By contrast, FedScale adopts a cost model to estimate the FL running time per round with a formula: $(\#Sample \cdot LatencyPerSample) + ModelSize/Bandwidth$. 
This estimation generates coarse-grained \textit{point} results and ignores factors that can directly affect the running time, such as I/O, hardware resource competition, and inaccurate pre-computation for $LatencyPerSample$ and $Bandwidth$.

\subsection{Scalability}
\label{exp:scale}
In the proposed \oursys, we enhance the concurrent processing capability of the FL server by providing multiple processes for Messages Transmitters and Processors (denoted as \textit{parallel FL transactions}), and providing process pools for parallel gRPC sending and receiving (denoted as \textit{parallel communications}), as introduced in Section \ref{sec:server}.
To confirm the effect of these concurrency techniques in improving the scalability of \oursys, we perform a stress test for the FL server, and compare the proposed \oursys with FedScale, FederatedScope (FS), and FS equipped only with \textit{parallel FL transactions}.
We design a stress generator in which the devices skip local training and only send/receive dummy model parameters, and track the average running time per FL round.

The experimental results are shown in Figure \ref{fig:our-scale}, from which we can conclude that the proposed \oursys performs much better than FS and FedScale in terms of scalability.
In particular, as the scales of clients increase, \oursys can effectively work in scenarios with up to 100,000 FL clients.
By contrast, when using the same high-performance server (1 TB memory and 64 CPU cores with 2.5 GHz frequency), FedScale takes 1.4x $\sim$ 3.9x time consumption longer than the proposed \oursys, and FS suffers from out-of-memory (OOM) at the scale of 100,000 and also excessive time consumption at the other scales.
Besides, we can see that when ablating the two concurrency optimizations (\ie, from red line to blue and green lines), the FL times significantly increase, which verifies the necessity and effectiveness of our optimizations that improve potentially heavy competition for different hardware resources and reduce the corresponding queuing delays.
In Appendix \ref{append:exp-scale-asyn}, we will further show more benefits of these two optimizations in asynchronous FL aggregation, which involves more intense competition for server system resources than the synchronized FL.

\section{How big is the gap w.r.t device runtime and scale?}
\label{sec:exp-gap}
In this section, we use the simulation platform provided in \oursys to investigate the FL performance in \textit{hetero-device} and \textit{scalable} scenarios, by varying the device distributions and the scales of participated clients.

\subsection{Simulation Settings}
\label{sec:exp-setting}
\subsubsection{Heterogeneous Devices.} 
We consider fairly diverse heterogeneous devices with different computation, memory, and communications capacities. Specifically, the devices can have 1$\sim$ 4 CPU cores, and CPU frequencies within 2.55 Ghz, 2.9 Ghz and 3.3 GHz. The device memory can be one of $\{256, 1024\}$ Mb. 
The network delays are one of $\{80 \sim 400,  35 \sim 200, 0\}$ in seconds, and the communication bandwidths (upload/download) can be one of $\{58,000/173,000,$ $75,000/285,000, $ $340,000/1,024,000\}$ in kbps, which are some representative configurations indicating network speeds and states of different quality such as 4G and WiFi.

\begin{figure}[h!]
	\centering
	\includegraphics[width=0.47\textwidth]{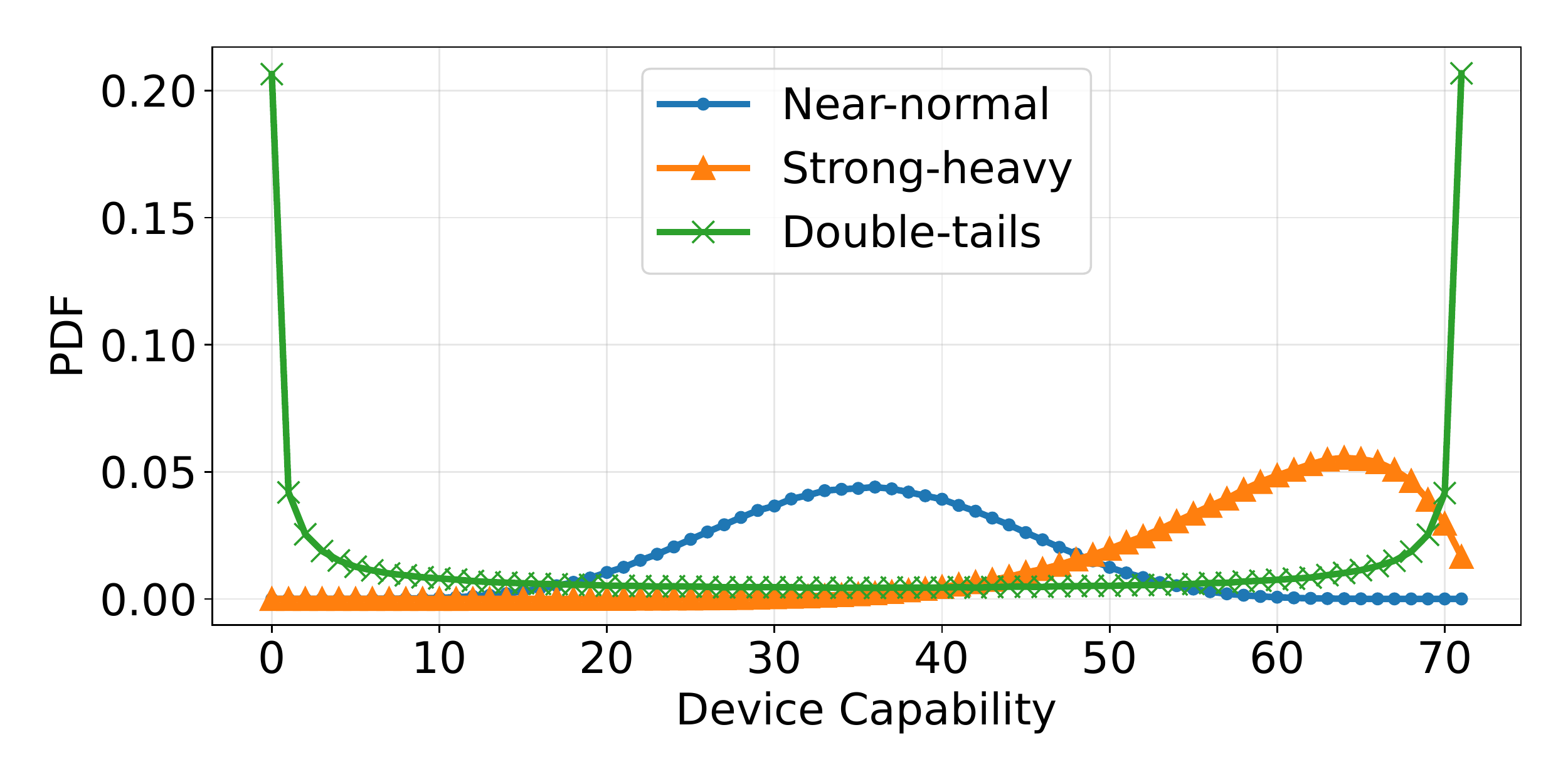}
        \caption{Illustration of different device distributions.}
        \label{fig:device-dist}	
\end{figure}

Based on the above hetero-device configurations, we consider generating multiple device distributions in the simulation device pool that corresponds to different application scenarios: 
(1) \textbf{Homo-device} case: all the participated devices have the same configuration (one of the configurations provided in \oursys), which is adopted by most existing works; 
(2) \textbf{Uniform} case: the devices have diverse configurations uniformly drawn from the set of provided configurations. This case differs \textit{homo-device} in only the heterogeneous device types;
(3) We further consider the device heterogeneity in terms of quantity (number of different device types). Specifically, we draw each device configuration from a beta-binomial distribution with parameters $(\alpha=10, \beta=10)$, which we denoted as \textbf{Near-normal} case corresponding to most applications where the devices with medium capacities have dominant numbers.
Similarly, we change the parameter of beta-binomial distribution into $(\alpha=10, \beta=2)$ for \textbf{Strong-heavy} case and $(\alpha=0.2, \beta=0.2)$ for \textbf{Double-tails} case where the major devices have strong and both strong and weak capacities respectively.
We illustrate the distributions in Figure \ref{fig:device-dist}, where the x-axis indicates the 72 device combinations ordered by their capacities (we sort the capacity matrix diagonally, and thus the larger index, the stronger capacity). 
Moreover, to simulate the changes in device runtime in real-world applications (\eg, CPU contention and network signal volatility), the device pool will randomly allocate available (\ie, no FL task is executing) devices to participated clients at each FL round.

\subsubsection{FL Settings.} 
We consider the widely adopted FL algorithm FedAvg and conduct experiments on the federated datasets FEMNIST, CelebA and Twitter \cite{leaf}.
Following previous works \cite{fedem,pflbench,fedbn}, we adopt the ConvNet model with different capacities for FEMNIST and CelebA datasets, and LR model for Twitter.
We vary the number of participated clients to investigate the effect of device scales $n$ while keeping the available device rate to be $0.3n$ at each FL round.
Due to the space limitation, we present more details about the implementation and adopted hyper-parameters in Appendix \ref{append:implent-detail}.

\subsubsection{Evaluation Metrics.}
We adopt comprehensive metrics to examine the FL performance in terms of  
(a) \textbf{Prediction Accuracy.} Both globally averaged and individual prediction accuracies are considered. Specifically, we calculate the average accuracy of each client weighted by their local data size (denoted as $\overline{acc}$),  the bottom 90\% decile (denoted as $\protect\widebreve{acc}$), and the standard deviation (denoted as $\sigma_{Acc}$) among all individual accuracies. 
(b) \textbf{Training and Communication Efficiency.} We track the FL convergence round and wall-clock times in hours (denoted as $T_{conv}$) to measure training efficiency, and use total communication bytes and network traffic (comm. bytes /sec) as metrics for communication efficiency.
(c) \textbf{Client Utilization.} For each participated client, we calculate the number of contributions per unit of time as $N_{contrib}/T_{conv}$, where $N_{contrib}$ indicates the total number of times the client successfully contributed to FL (uploaded models for aggregation). We consider the mean (denoted as $\overline{uti}$) and standard deviation (denoted as $\sigma_{uti}$) for this metric among all clients to reflect the extent to which this federation unites the various participants. 
All the experiments are repeated 2 times with different random seeds.

\begin{figure*}[h!]
	\centering
	
	\subfigure[$\overline{acc}$]{
		\includegraphics[width=0.31\textwidth]{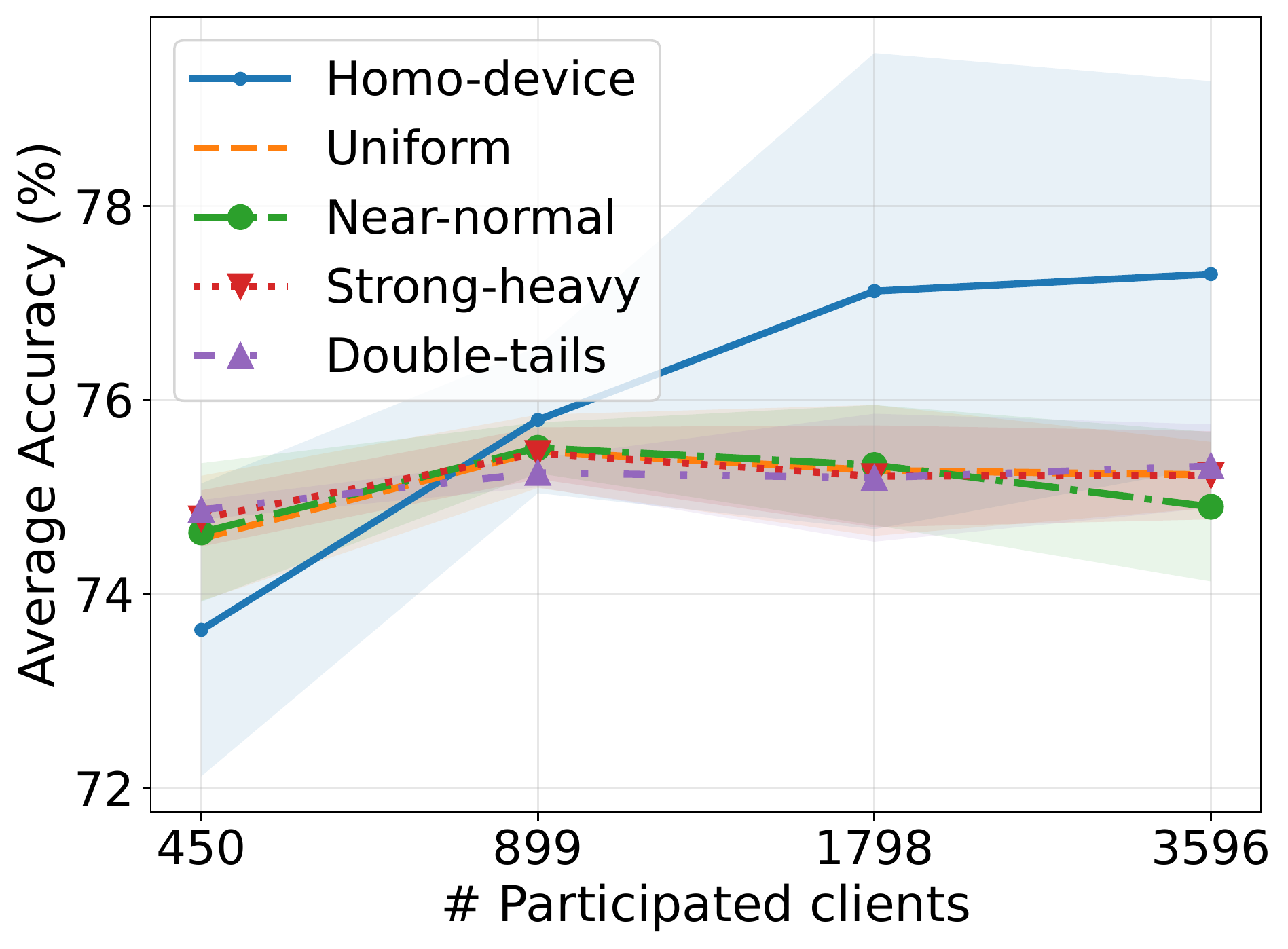}
		\label{fig:acc-parti}
	}
	\subfigure[$\protect\widebreve{acc}$ ]{
		\includegraphics[width=0.31\textwidth]{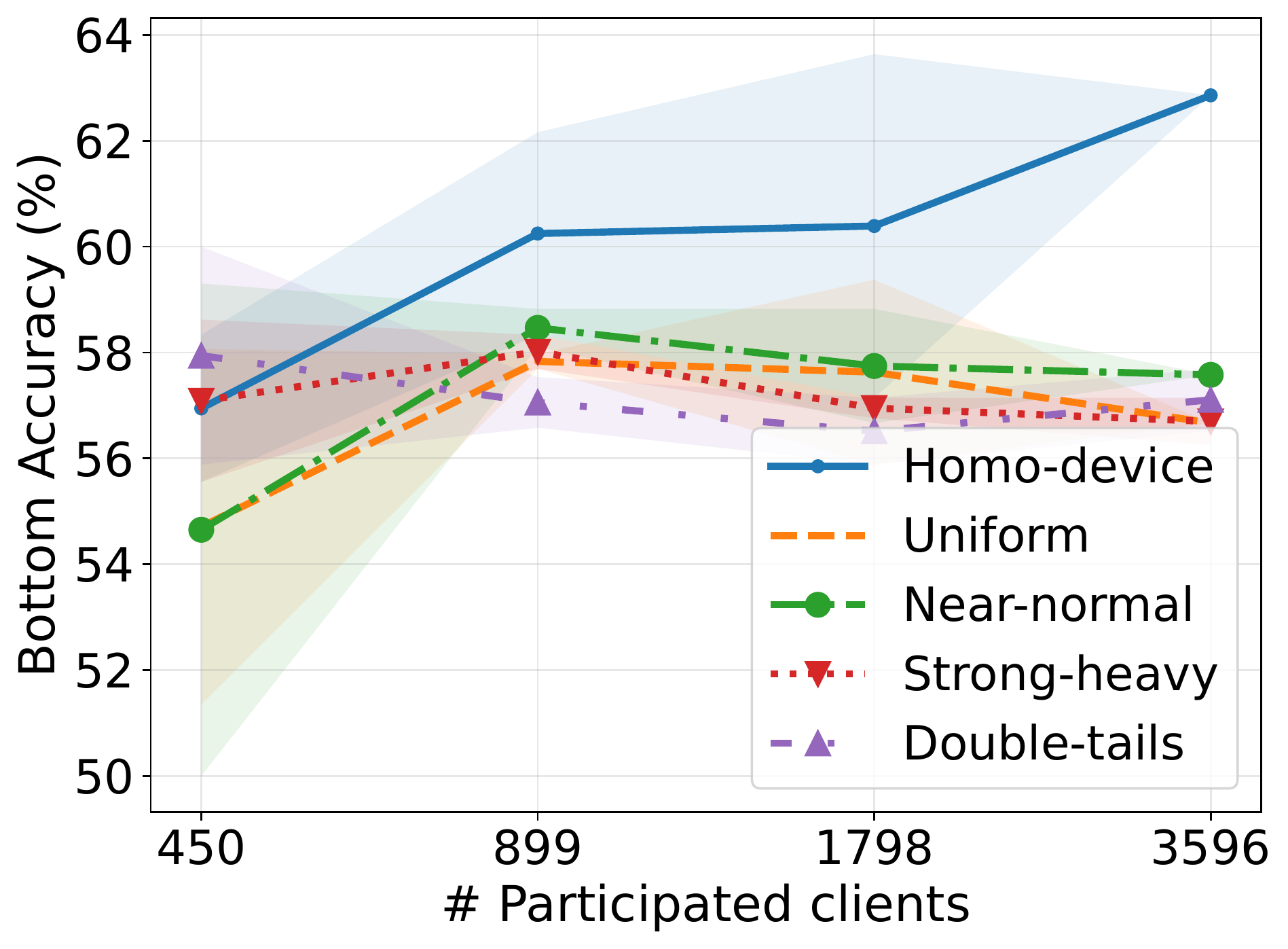}
	}
	\subfigure[ $\sigma_{Acc}$]{
		\includegraphics[width=0.31\textwidth]{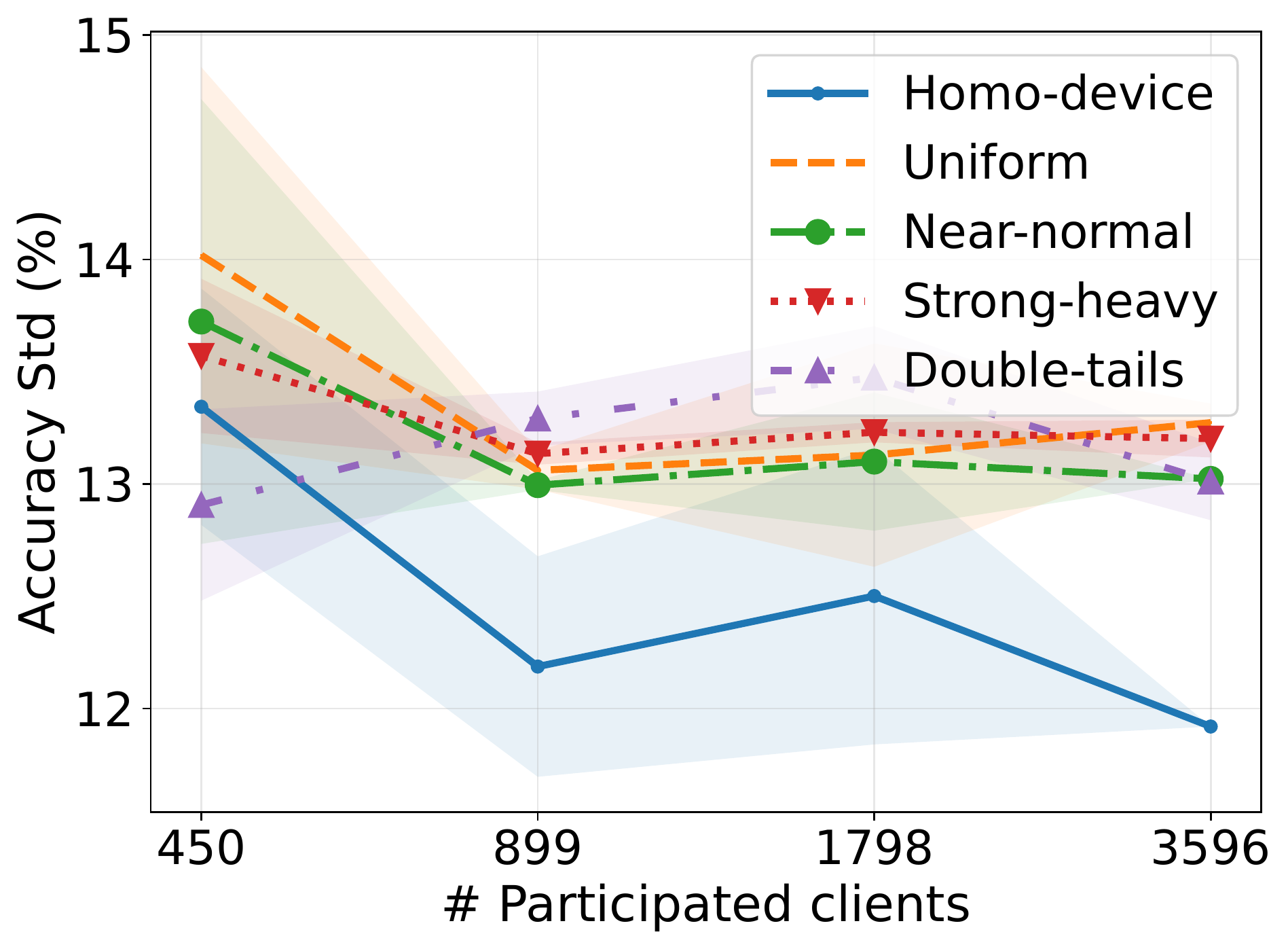}
		\label{fig:std-acc-parti}
	}
	\caption{Accuracy w.r.t. increasing number of \textit{participated} clients under different hetero-device distributions on FEMNIST. }
	\label{fig:acc-scale-curve-parti}
\end{figure*}

\begin{figure*}[h!]
	\centering
	\subfigure[$\overline{acc}$]{
		\includegraphics[width=0.31\textwidth]{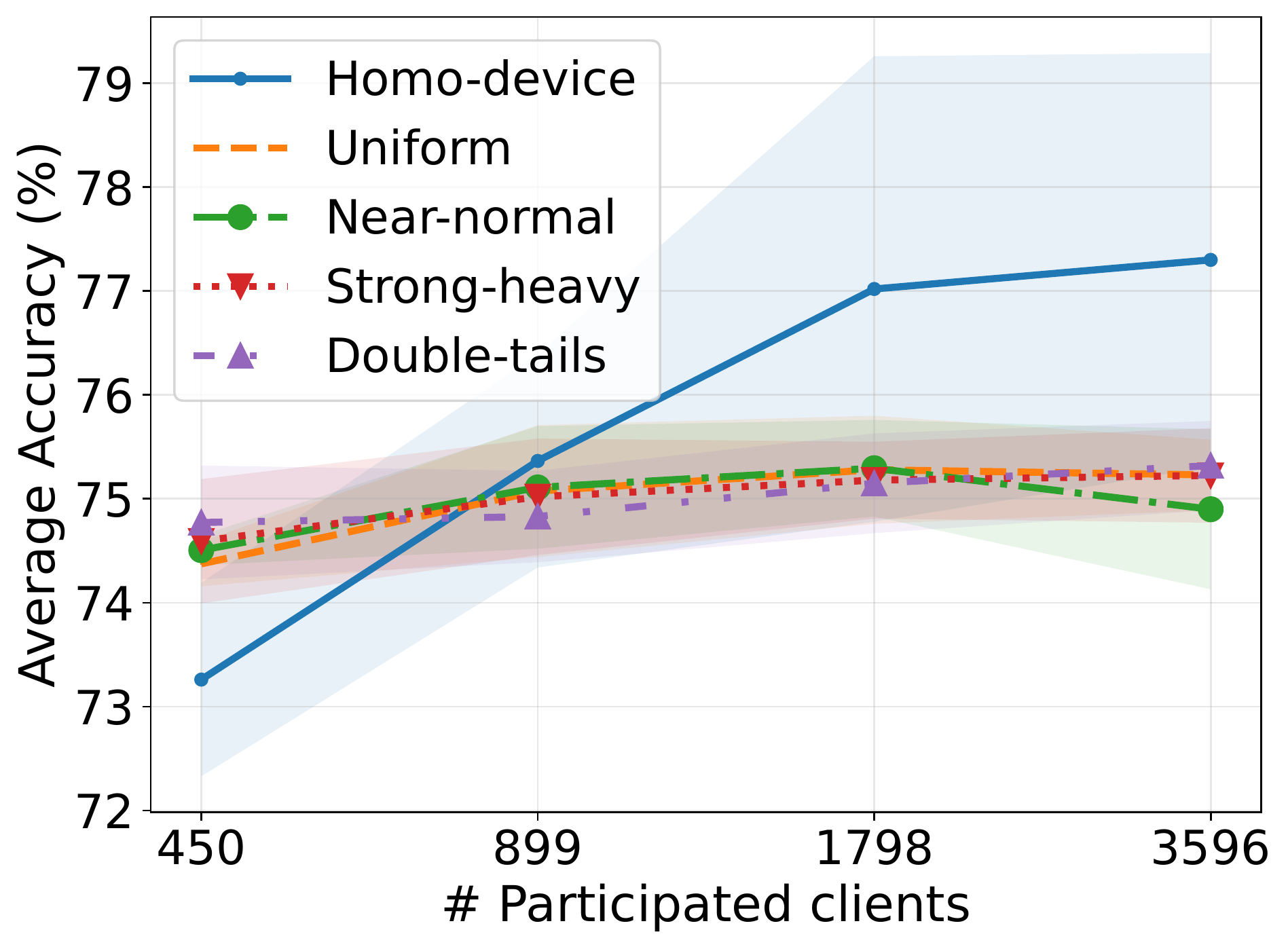}
		\label{fig:acc-all}
	}
	\subfigure[$\protect\widebreve{acc}$ ]{
		\includegraphics[width=0.31\textwidth]{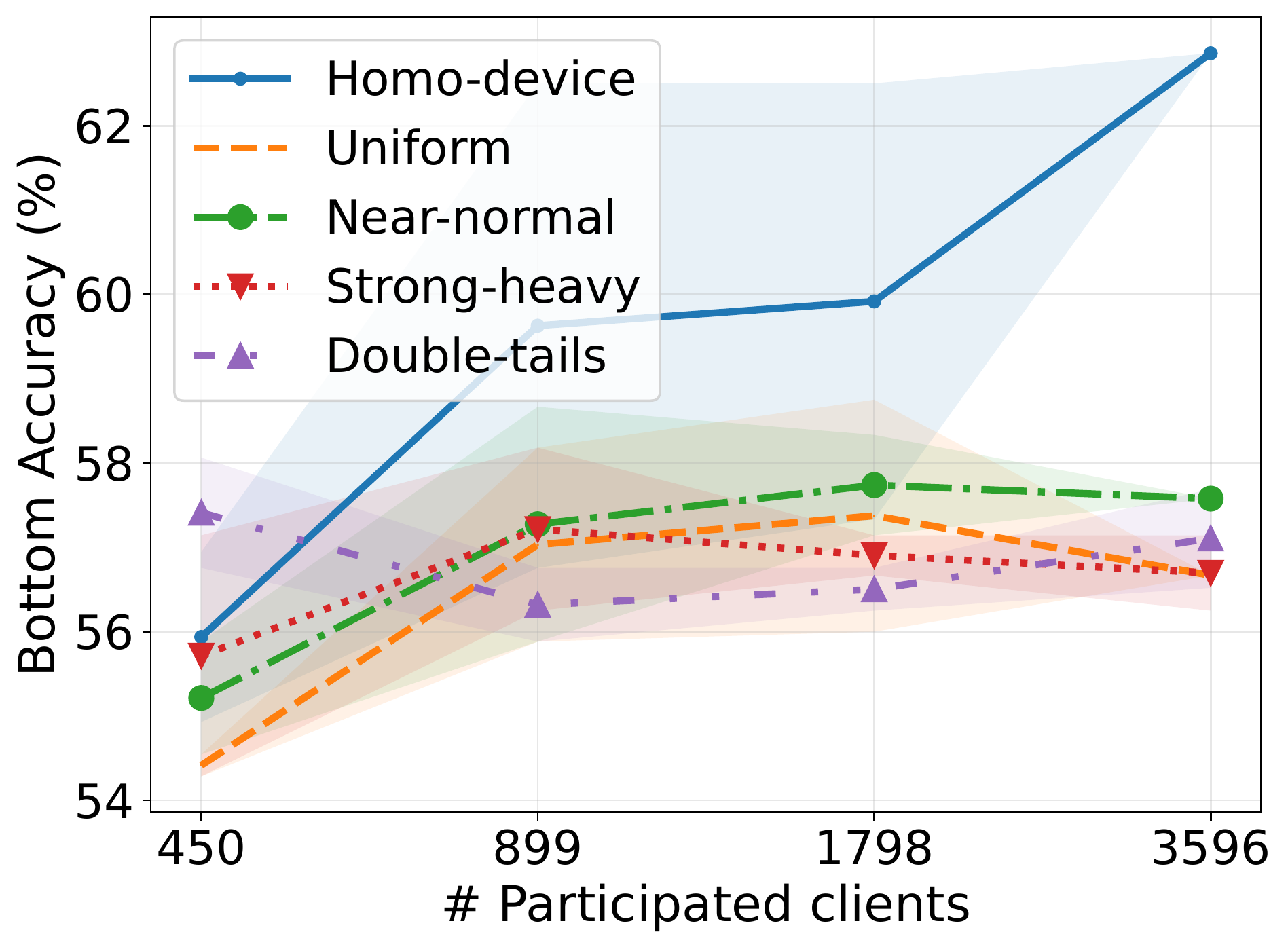}
		\label{fig:bottom-acc-full}
	}
	\subfigure[ $\sigma_{Acc}$]{
		\includegraphics[width=0.31\textwidth]{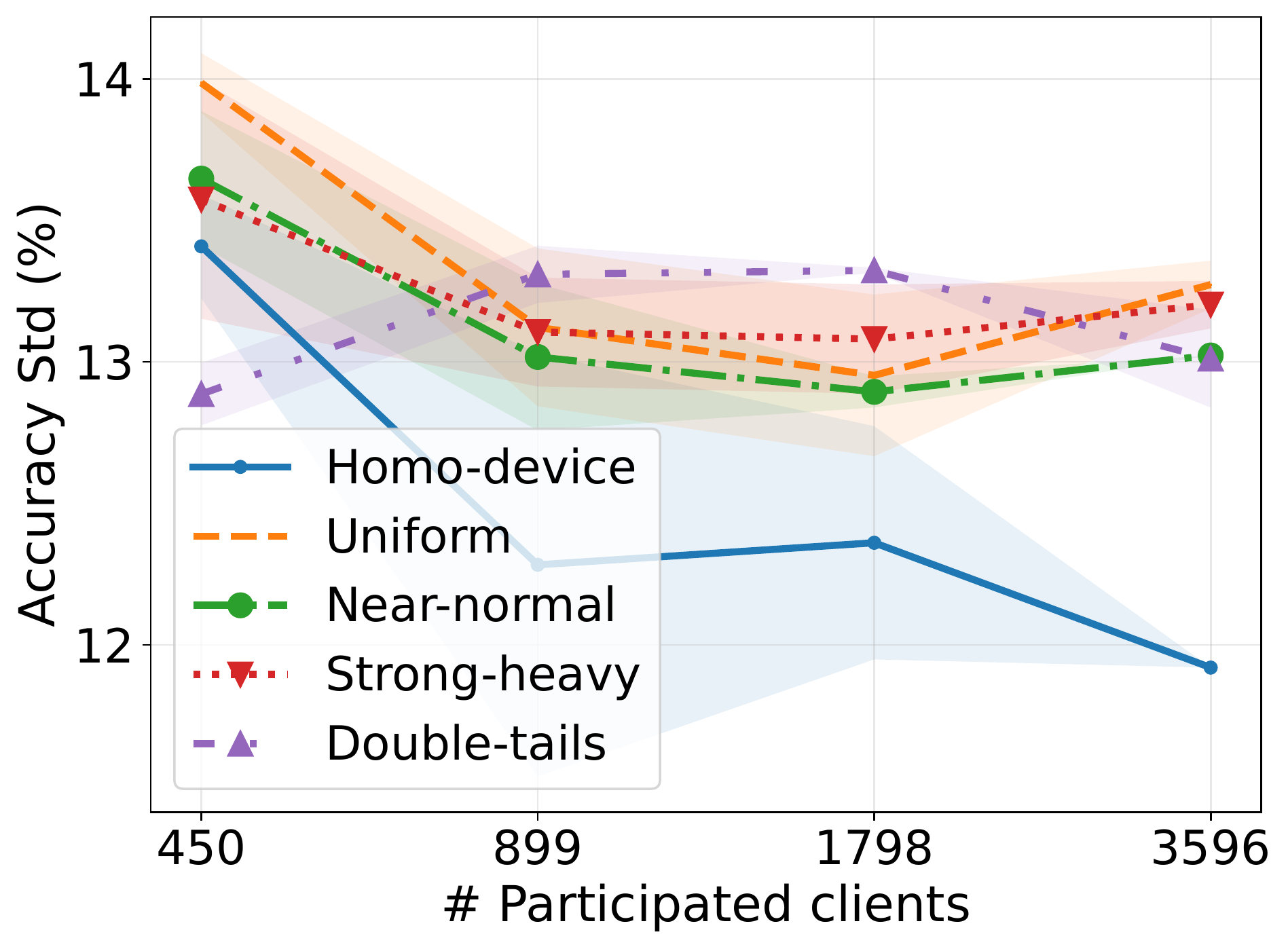}
	}
	\caption{Accuracy w.r.t. increasing number of \textit{all} clients under different hetero-device distributions on FEMNIST.}
	\label{fig:acc-scale-curve-full}
\end{figure*}

\subsection{Accuracy and Fairness}
We show various accuracy results ($\overline{acc}, \protect\widebreve{acc}$ and $\sigma_{acc}$) under different hetero-device distributions on FEMNIST in Figure \ref{fig:acc-scale-curve-parti} and Figure \ref{fig:acc-scale-curve-full}. The results within Figure \ref{fig:acc-scale-curve-parti} are evaluated on the participated clients (\eg, the number of evaluated clients is 450, 900, $...$), while the results within Figure \ref{fig:acc-scale-curve-full} are evaluated on all clients of the full datasets (\ie, the 3,596 clients of FEMNIST).

From these figures, we can see that
\textbf{there is a substantial accuracy gap between the homo-device case and hetero-device cases, which is more noticeable in the fairness-related metrics and at large device scales.}
Specifically, for both the participated and full clients (Figure \ref{fig:acc-parti} and Figure \ref{fig:acc-all}), the $\overline{acc}$ differences between homo- and hetero-device cases are not negligible, and the difference increases as the clients scale increases (from $-$1.6\% at the scale of 450 to 2.7\% at the scale of 3,596). 
When considering the fairness-related metrics, the differences become larger, \eg, at the scale of 3,596, there is about $7\%  \protect\widebreve{acc}$ gap in Figure \ref{fig:bottom-acc-full} and about $1.5\%$ ${\sigma}_{acc}$ gap in Figure \ref{fig:std-acc-parti}, this is because that the inconsistent training dynamics between homo- and hetero-device cases have a greater impact on individual clients, especially on those having slow response speeds.
We also find that with the increasing number of participated clients, the accuracy of homo-device setting improves by about $3\%\sim4\%$ for $\overline{acc}$, and  $6\%\sim7\%$ for $\protect\widebreve{acc}$.
However, the results of other hetero-device distributions have much smaller changes,
leaving the huge potential to improve the accuracy and practicality of FL under hetero-device and scalable scenarios.

Moreover, \textbf{both distribution differences and quantity differences in heterogeneous devices do matter.} 
Among the compared hetero-device distributions, there are large homo-hetero differences of $\overline{acc}$, $\protect\widebreve{acc}$ and ${acc}_{\sigma}$ even for the \textit{Uniform} distribution case, which differs \textit{Homo-device} case in only the participated device types.
When taking the quantity differences into consideration, the accuracy differences become larger for the other hetero-device cases, such as $2.7\%$ for $\overline{acc}$ metric on \textit{Near-normal} distribution at scale of 3,596.
Besides, the variances of results of homo-device w.r.t. different FL experiments are much larger than the ones of hetero-device cases due to the fact that all clients have the same opportunity to contribute to FL aggregation in homo-device distribution, verifying the existence and significance of the gap again.

\begin{figure*}[h!]
	\centering
	\includegraphics[width=0.31\textwidth]{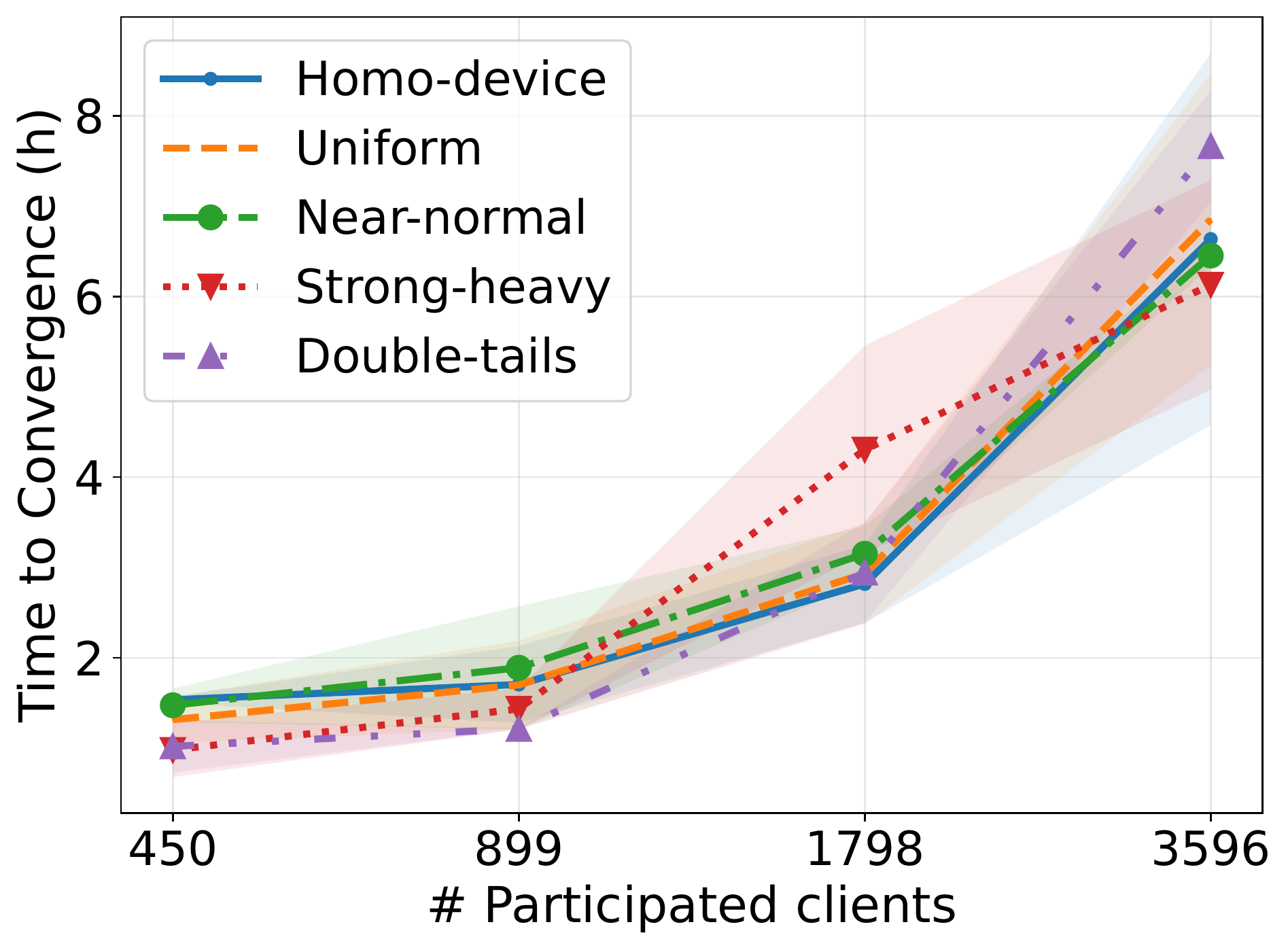}
	\includegraphics[width=0.31\textwidth]{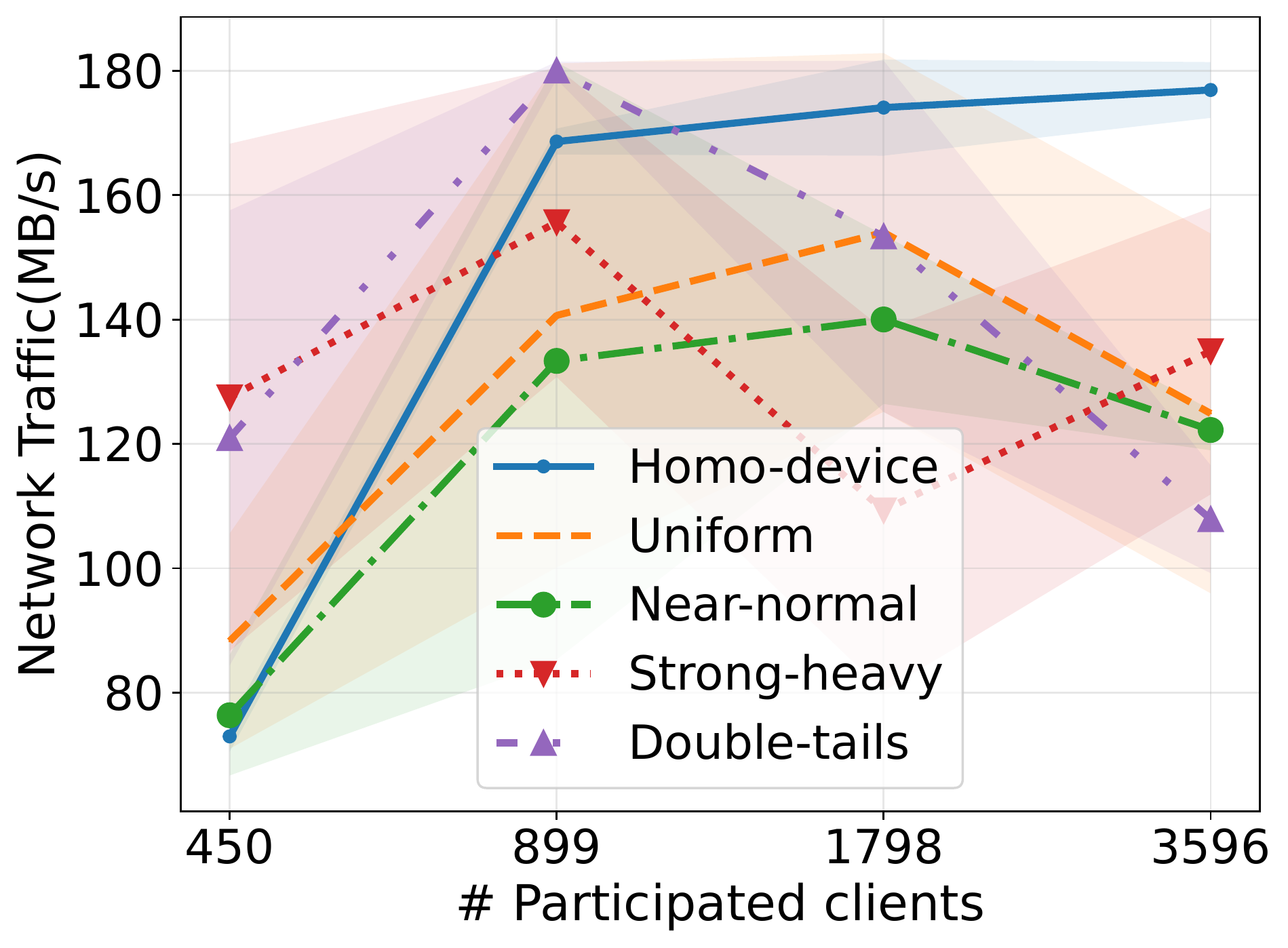}
	\includegraphics[width=0.31\textwidth]{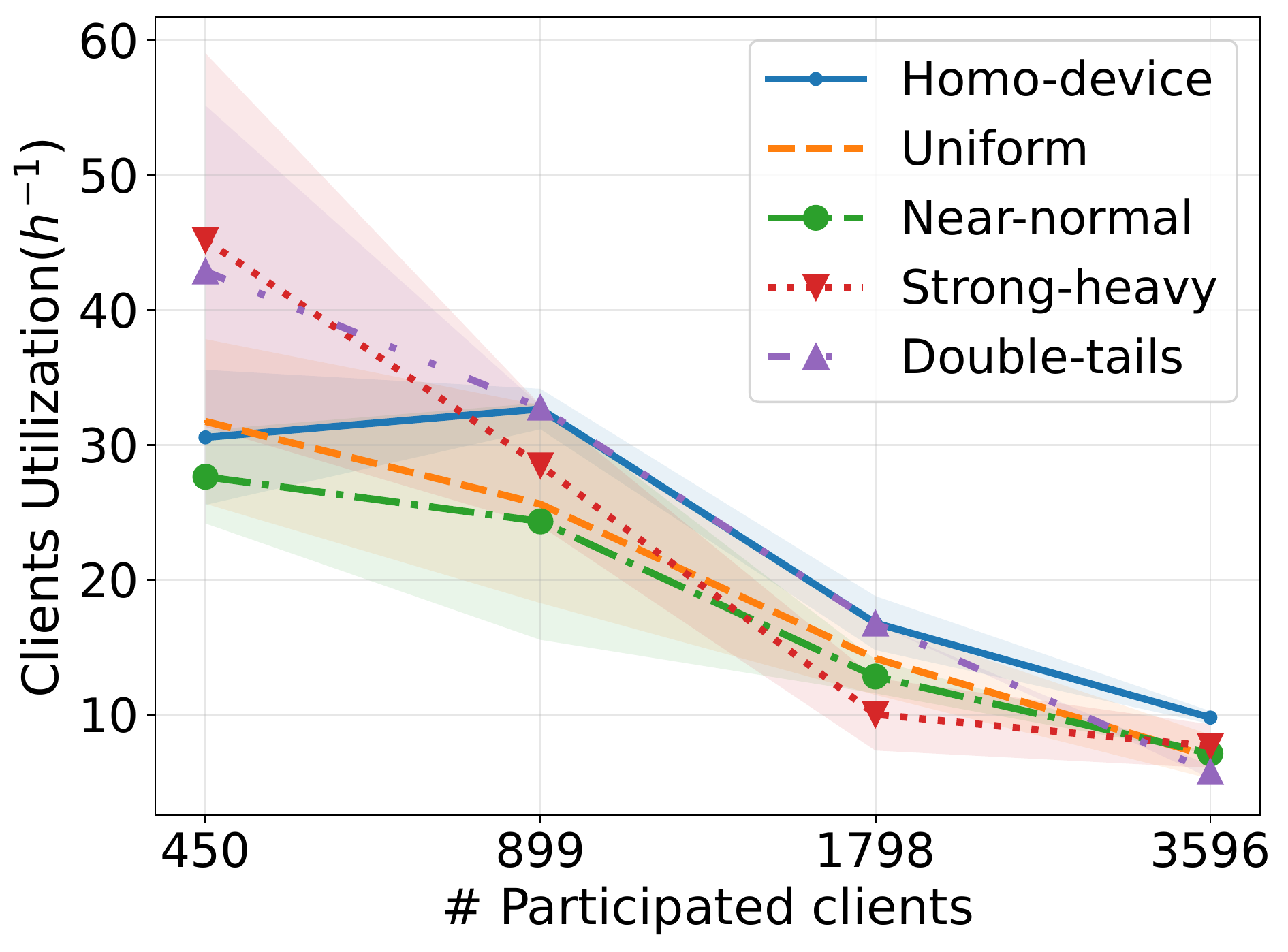}
	\caption{Performance under different hetero-device distribution w.r.t time to convergence (left), communication traffic (middle), and clients utilization (right) on FEMNIST.}
	\label{fig:sys-curve}
\end{figure*}

\begin{table*}[h!]
\centering
\caption{Experimental results of personalized FL algorithms on FEMNIST under \textit{\textit{Near-normal}} device distribution.}
\label{tab:pfl-study}
\begin{tabular}{lcccccccccccc} 
\toprule
\multirow{2}{*}{Method } & \multicolumn{4}{c}{\# Clients = 225} & \multicolumn{4}{c}{\# Clients = 450} & \multicolumn{4}{c}{\# Clients = 900} \\
\cmidrule(lr){2-5}  \cmidrule(lr){6-9}  \cmidrule(lr){10-13}
& $\overline{acc}$~ & $\protect\widebreve{acc}$ & $\sigma_{acc}$ & $T_{conv}$ & $\overline{acc}$ & $\protect\widebreve{acc}$  & $\sigma_{acc}$ & $T_{conv}$ & $\overline{acc}$ & $\protect\widebreve{acc}$  & $\sigma_{acc}$ & $T_{conv}$ \\ 
\midrule
FedAvg & 70.21 &  54.23 &   13.17   & 0.42 & 74.36 &    55.21   &   13.65   & 1.47 & 75.70 &  57.27   &  13.01    & 1.89 \\ 
FT & 72.62 &  54.05 &  16.24    & 0.76 & 75.16 &  58.82  &  15.05  & 1.72 & 75.36 &  59.38 &  15.22    & 1.96 \\ 
FedBABU & 74.03 &  55.39    &  18.43    & 0.97 & 75.01 & 58.57   &  15.95    & 1.16 & 75.83 &  58.29   &  17.15    & 1.48 \\
\bottomrule
\end{tabular}
\end{table*}

\subsection{FL System Efficiency}
In addition to model accuracy, we also quantify the impact of device heterogeneity and scales on the system efficiency.
Figure \ref{fig:sys-curve} demonstrates the results of the convergence wall-clock time in hours, the network traffic (communication bytes per second), and the client utilization (the contribution number per hour averaged over all clients, $\overline{uti} \pm \sigma_{uti}$).
Compared with the results for homo-device case, we observed that hetero-device cases exhibit \textbf{complex differences in the time to convergence, network traffic, and client utilization specific to device scales and device distributions}, challenging the utility of FL in real-world scenarios involving constrained time resources, constrained network traffic, and user incentive mechanisms.

Specifically, 
(1) The variance of time to convergence increases with increasing number of participated clients (e.g. $0.09$ with 450 participated clients and $1.34$ with 3,596 participated clients for \textit{Strong-heavy} case). 
On the one hand, this instability of convergence rate affects the iteration scheduling of real FL applications, as well as potentially more resource usage than budgeted. 
On the other hand, the degree of fluctuation is very different across heterogeneous distributions and scales, suggesting that different hetero-device distributions and scales amplify the differences in convergence due to the initialization of the model parameters and the aggregation dynamics (we varied the random seeds across multiple experiments), leaving an open question about how to design FL algorithms with more stable convergence under real FL scenarios.

(2) For communication load, interestingly, hetero-device cases have on average lower communication loads than homo-device case (e.g., -13.1\% for \textit{Strong-heavy} case) and on average significantly larger load variance than homo-device case (+4,989\%). 
This indicates that in real FL applications, the network traffic consumed by the entire FL server (the FL service provider) and clients (probably the individual users) in an acceptable time may be less than the estimated results with homogeneous devices. Also note that our system enables timeout re-broadcasting and over-selection mechanisms to match the real scenarios, thus some wasted traffic is also included in the reported traffic load. How to further reduce traffic waste is still under-explored, especially in conjunction with the heterogeneous device distributions and scales.

(3) For client utilization, except for the two highly biased hetero- distributions (\textit{Strong-heavy} and \textit{Double-tails}), the average utilization of the other ones is lower than that of the homo-device case. 
Besides, the utilization variance under hetero-device cases are significantly larger than those of the homo-device scenarios, especially with medium device scales, which implies that more clients are not able to contribute their uploaded messages efficiently to FL aggregation. 
It is important to improve the $\overline{uti}$ and reduce the $\sigma_{uti}$ in scalable hetero-device FL that involve incentive mechanisms \cite{zhan2020learning,zeng2021comprehensive}.

\begin{table*}[h!]
	\centering
	\caption{Experimental results of personalized FL algorithms on CelebA under \textit{\textit{Near-normal}} device distribution.}
	\label{tab:pfl-study-celeba}
\begin{tabular}{lcccccccccccc} 
\toprule
\multirow{2}{*}{Method} & \multicolumn{4}{c}{\# Clients = 250} & \multicolumn{4}{c}{\# Clients = 500} & \multicolumn{4}{c}{\# Clients = 1000} \\
\cmidrule(lr){2-5}  \cmidrule(lr){6-9}  \cmidrule(lr){10-13}
 & $\overline{acc}$~ & $\protect\widebreve{acc}$ & $\sigma_{acc}$ & $T_{conv}$ & $\overline{acc}$ & $\protect\widebreve{acc}$  & $\sigma_{acc}$ & $T_{conv}$ & $\overline{acc}$ & $\protect\widebreve{acc}$  & $\sigma_{acc}$ & $T_{conv}$ \\ 
\midrule
FedAvg &  74.57 &   50.00   &  23.56 & 0.73 & 73.31 & 33.33 & 25.28 & 1.36 & 75.36 & 50.00 & 22.49 & 1.68 \\ 
FT &  76.74    &    50.00   &   23.32   &  0.78   & 79.35 &    50.00   &   22.49  &   1.36   & 79.66 &  50.00  &   21.54  &  1.70 \\ 

FedBABU & 77.21 & 40.00 & 24.86 & 0.77 & 79.20 & 50.00 & 21.39 & 1.44 & 80.64 & 50.00 & 25.01 & 1.73 \\
\bottomrule
\end{tabular}
\end{table*}

\begin{table*}[h!]
	\centering
	\caption{Experimental results of compression techniques on FEMNIST under \textit{\textit{Near-normal}} device distribution.}
	\label{tab:compress-study}
	\begin{tabular}{lccccccccc}
		\toprule                      
		\multirow{2}{*}{Method} & \multicolumn{3}{c}{\# Clients = 225} & \multicolumn{3}{c}{\# Clients = 450} & \multicolumn{3}{c}{\# Clients = 900}  \\
		\cmidrule(lr){2-4}  \cmidrule(lr){5-7}  \cmidrule(lr){8-10}
		& $\overline{acc}$  & net. traffic & $T_{conv}$  &   $\overline{acc}$     & net. traffic &   $T_{conv}$ &  $\overline{acc}$ &   net. traffic &  $T_{conv}$  \\ \midrule
		FedAvg w/o comp.    &   70.21  &    135.08  &   0.42    &   74.36  &   106.02   &   1.47     &  75.70   &  133.34   &   1.89    \\ 
		+Gzip &   70.43  &   90.24   &   0.54   &  73.75  &   93.78 &  0.98 &   75.36  &    94.62  &   2.12    \\ 
		+FP16  &   69.63    &   56.23 &   0.44    &   73.99 &   64.38    &   0.78      &   74.58   & 88.71   &   1.16      \\ 
		+INT8   &   65.58  &   32.39   &   0.41    &   72.63   &   42.58  &   0.63   &   74.46   &   43.29   &   1.19     \\  
		\bottomrule		
	\end{tabular}
\end{table*}

\section{Advanced FL Features with \oursys}
\label{exp:advanced-fl-tech}
In this section, we will show several use cases of advanced FL features with \oursys, including personalization (Section \ref{exp:pfl}), compression (Section \ref{exp:compression}), and asynchronous training (Section \ref{exp:asyn}). Due to the space limitation, we present more results about the performance gap between homo- and hetero- device distributions for these studied advanced FL features in Appendix \ref{append:homo-hetero-advanced-fl}, and more results about the end-to-end evaluation when combining these studied advanced FL features in Appendix \ref{append:end2end}.
\subsection{Personalized FL (pFL)}
\label{exp:pfl}
Personalization is one of the promising directions of FL to maximize the model utility for individual clients.
To demonstrate how well \oursys supports personalized FL research, we take FedAvg as a baseline and compare it with fine-tuning (FedAvg + FT) and FedBABU \cite{oh2022fedbabu}, a SOTA pFL algorithm.
Specifically, FedBABU freezes the final classification layer during FL training, lets clients train and upload other layers of the model to the server for aggregation, and fine-tune the whole model before evaluation.
With the results of FedBABU on \oursys, we want to demonstrate the potential of \oursys for supporting more customized training/upload/aggregation behaviors in the FL personalization study.

We summarize the results in Table \ref{tab:pfl-study} and \ref{tab:pfl-study-celeba} based on heterogeneous devices following \textit{\textit{Near-normal}} distribution for FEMNIST and CelebA dataset respectively.
In general, the pFL algorithms achieve better accuracy than FedAvg (both $\overline{acc}$ and $\protect\widebreve{acc}$) on FEMNIST dataset, while larger $\sigma_{acc}$ with un-fairness.
Besides, the performance advantages of pFL algorithms over FedAvg decrease on FEMNIST as the device scale increases (\eg the $\overline{acc}$ difference between FedBABU and FedAvg is 3.82, 0.65 and 0.13 at the scale of 225, 450, 900 respectively). 
Interestingly, compared to the results on FEMNIST, the accuracy advantage of pFL algorithms over FedAvg is more significant on CelebA dataset (\eg, the absolute $\overline{acc}$ improvements are 2.64\% and 5.89\% on FEMNIST and CelebA respectively).
This discrepancy may be due to the different amount of local data (the average number of local data size is 21.4 for CelebA, which is less than FEMNIST's 226.8).
It is suggested to improve the fairness and stability for future personalized FL algorithms by collaboratively accounting for device heterogeneity and data heterogeneity.

\begin{table*}[h!]
	\centering
	\caption{Experimental results of asynchronous aggregation on FEMNIST under different device distributions.}
	\label{tab:asyn-study}
	\begin{tabular}{lcccccccccccc}
		\toprule                       
		\multirow{2}{*}{Method} & \multicolumn{4}{c}{\# Clients = 225} & \multicolumn{4}{c}{\# Clients = 450} & \multicolumn{4}{c}{\# Clients = 900}  \\
		\cmidrule(lr){2-5}  \cmidrule(lr){6-9}  \cmidrule(lr){10-13}
		& $\overline{acc}$ & $\protect\widebreve{acc}$ & $\sigma_{acc}$ & $\overline{uti} \pm  \sigma_{uti}$ &   $\overline{acc}$ & $\protect\widebreve{acc}$ & $\sigma_{acc}$ &       $\overline{uti} \pm  \sigma_{uti}$  &  $\overline{acc}$ & $\protect\widebreve{acc}$ & $\sigma_{acc}$ &       $\overline{uti} \pm  \sigma_{uti}$     \\ \midrule
		Sync, \textit{Near-normal}    &   70.21  &    54.23  & 13.17  &   41.68$\pm$18.87&   74.36  &   55.21  & 13.65  &   27.63$\pm$3.98    &  75.70   &  57.27  & 13.01   &   24.33$\pm$3.72   \\ 
		Async, \textit{Near-normal}        &   80.81  &  67.65  & 10.01   &   54.93$\pm$3.81  &   77.97   & 62.50  & 11.77  &   72.84$\pm$9.14    &   78.34   & 63.01 &  14.57  &   79.41$\pm$4.31    \\ \midrule
		Sync, \textit{Strong-heavy}  &   70.74   & 54.99  &  12.66  &   46.90$\pm$12.49 &   74.23  &  57.41 & 12.89  & 42.86$\pm$6.94   &   74.83  & 56.32 &  13.31 &  32.75$\pm$5.34   \\ 
		Async, \textit{Strong-heavy}        &   77.84  & 55.88  &  14.20  &   59.02$\pm$4.70  &  73.43  &  54.55   &  13.20 &   77.16$\pm$8.69   &    74.94  &  56.25  &  16.26  &   81.32$\pm$6.95                   \\
		\bottomrule
		
	\end{tabular}
\end{table*}

\subsection{Communication Compression}
\label{exp:compression}
In real FL scenarios, communication cost is one of the most important metrics, especially in low-resource cross-device federated learning. 
To illustrate how the compression techniques affect the trade-off between the communication cost and model utility on \oursys, we conduct experiments to compare accuracies, communication cost and time to convergence of the FedAvg using vanilla gradients transmission (no compression), transmission with Gzip (lossless compression), quantized transmission in FP16 and INT8 (lossy model quantization).
The results on FEMNIST are summarized in Table \ref{tab:compress-study}.

One can see from Table \ref{tab:compress-study} that, as the number of clients increases, the accuracies of models using different communication modes also increase as expected for the \textit{\textit{Near-normal}} distribution.
The reducing network traffic effect is significant when the compression communication mode is turned on. 
Compared with the communication cost of the vanilla communication mode, Gzip reduces the 33\% cost with almost indistinguishable accuracy differences.
As for the lossy compressions, FP16 reduces about 58\% communication cost while INT8 has only 1/3 communication cost of the vanilla one.
Although lossy compression introduces slight accuracy declines, the gap between the vanilla communication mode and the lossy ones is narrowed down to only about 1\% differences when the number of clients increases to 900.
Another benefit of using lossy compression lies in the time to convergence of FL training.
We also measure the convergence time (wall time in hours) and show that the training with lossy compression usually takes less time to converge.

Overall, by comparing the result of different communication modes, we demonstrate that our \oursys can have great potential in federated learning model compression.

\subsection{Asynchronous Aggregation}
\label{exp:asyn}

Applying asynchronous aggregation in FL balances the training efficiency and model utility, since clients get out of waiting for stragglers to finish local training at each FL round.
For cross-device FL, asynchronous aggregation is particularly essential for handling the heterogeneity, however, it also brings additional challenges to the system design.
For example, compared with synchronous aggregation, applying asynchronous aggregation might create more serious resource competition for sending and receiving messages, and lead to the training time growing rapidly with the scale.
With the help of the FL Server in \oursys (Section~\ref{sec:server}), separate process pools are provided for parallel sending and receiving messages, which makes it scalable and efficient to apply asynchronous aggregation in FL.

Furthermore, we provide observations and discussions on the effectiveness and efficiency of asynchronous aggregation with different device distributions and device scales, calling for further careful adaptation towards real-world hetero-device FL.
The experimental results are reported in Table~\ref{tab:asyn-study}. Overall, we can observe that applying asynchronous aggregation achieves competitive model performance and significantly higher utilization of clients' contributions compared to those of synchronous aggregation.
When the device scale increases,  the utilization becomes higher when applying asynchronous aggregation (e.g., 59.02/77.16/81.32 for 225/450/900 participated clients under \textit{Near-normal} device distributions), while the utilization becomes lower when applying synchronous aggregation (e.g., 46.90/42.86/32.75 for 225/450/900 participated clients under \textit{Near-normal} device distributions). These results confirm the importance and necessity of asynchronous techniques towards real-world hetero-device FL.
Besides, it is worth pointing out the impact of different device distributions on the model performance (e.g., $\overline{acc}$ and $\sigma_{acc}$). When using synchronous aggregation, the model performance varies slightly (e.g., $\pm$0.53/0.13/0.87$\%$ on $\overline{acc}$ when the number of clients is 225/450/900 comparing between \textit{Near-normal} and \textit{Strong-heavy}) under different device distributions. However, there exist noticeable variances (e.g., $\pm$2.97/4.54/3.40$\%$ on $\overline{acc}$ when the number of clients is 225/450/900 comparing between \textit{Near-normal} and \textit{Strong-heavy}) in model performance when applying asynchronous aggregation under different device distributions.
These variances inspire us to call on the community to pay more attention to the hetero-device distribution of real-world FL applications for improving the usability and robustness of cross-device FL studies, and to carefully adapt some async-mode configurations (e.g., staleness toleration and aggregation goal) and develop new FL algorithms according to the hetero-device distributions.

\section{Conclusions}
While most existing FL studies consider little about the heterogeneous devices with various scales in practice, in this paper, we introduce our released system \oursys, which is featured by its high usability and strong robustness for handling real-world cross-device FL tasks with a massive amount of heterogeneous client devices.
This is achieved by the key components of the system, including the optimized device executor for efficient and diverse mobile runtimes, and the enhanced FL server for scalable message transmission and transaction execution.
In addition, our system provides flexible programming interfaces for several advanced FL topics, including personalization, compression and asynchronous aggregation.
We further implement a high-fidelity heterogeneous device simulation platform to give researchers and developers an easy-to-run option for cost-efficiently prototyping FL solutions. 
All the above contributions are demonstrated via our extensive experiments, in which we also investigate the impact of device heterogeneity and device scales for real-world FL tasks.
We hope that our system can facilitate further real-world FL studies and applications, and our observations and discussions can contribute to a deeper understanding of the impact of device heterogeneity and device scales in the FL community.

\bibliographystyle{ACM-Reference-Format}
\bibliography{fsdevice}

\clearpage
\appendix

\section*{Appendix}

\section{Implementation Details}
\label{append:implent-detail}

\subsection{Datasets}

Our experiments are conducted on several widely used FL datasets with diverse scales and tasks.
Specifically, \textbf{FEMNIST} \cite{leaf,emnist} is for hand-written digits classification and contains 3,550 clients whose local data is partitioned by the writers. \textbf{CelebA} \cite{celeba,leaf} is for the classification of celebrities' characteristics and contains 9,343 clients whose local data is partitioned by the celebrities. \textbf{Twitter} \cite{twitter,leaf} is for sentiment classification and we adopt the subset used in \cite{pflbench}, which contains 13,203 clients whose local data is partitioned by the Twitter users. 
We randomly split the datasets into train/valid/test sets with a ratio of 6:2:2.

\subsection{Models and Baselines}
Following previous works \cite{leaf,federatedscope,fedem,pflbench}, we adopt CNN models for FEMNIST and CelebA datasets, and an LR model for the Twitter dataset.
Specifically, 
for FEMNIST, the training model consists of two convolution layers and two linear layers, whose dimensions are 32, 64, 1024 and 62 respectively.
For CelebA, we use the model with the same architecture as the one used for FEMNIST, but with dimensions [32, 64, 256, 2] to avoid out-of-memory in devices.
For Twitter, we use an LR model and represent the sentence features by concatenating the 50d Glove embeddings. \footnote{https://nlp.stanford.edu/data/glove.6B.zip.}

\subsection{Platform and Hyper-Parameters}
We conduct the experiments on a cluster of 10 servers, whose CPU cores are within [196, 256, 512] and the CPU frequencies are within [2.55, 2.9, 3.3] Ghz.
The codes are built upon FederatedScope in version 0.2.0, MNN in version 2.0.0, Clang in version 14.0.0, and android JDK in version 1.8.

For each adopted dataset, we vary the number of participated clients to investigate the effect of device scales $n$, while keeping the available device rate to be $0.3n$ at each FL round, i.e., the server only sends messages to $0.3n$ randomly selected clients at each FL round.
As we mentioned in Section \ref{sec:server}, there are slow and disconnected devices in real hetero-device scenarios, we thus enable the over-selection with ratio $q=1.5$ for all experiments in synchronous mode, and enable the timeout mechanism with the timeout parameters $t_o=60, \delta_t=5$.
Besides, we run FL in at most $200$ rounds, set the local run step to be 1 epoch for each device, and grid search the hyper-parameters of FL algorithms.

In synchronous aggregation, the learning rate is searched from $\{0.001,0.025, 0.05, 0.1,0.5, 1\}$. We adopt 0.1 for FEMNIST and 0.025 for CelebA. 
We train FL models for 200 training rounds with a batch size of 16 for both datasets.
In Section \ref{exp:pfl}, we locally fine-tune the models for 5 epochs for each client before evaluation. 
In asynchronous aggregation, similarly, we train the FL models for 200 training rounds with a local learning rate of 0.1. The batch size is set as 16.

\section{Additional Experiments}
\label{append:more-exp-results}

\subsection{Scalability Study for Asynchronous Aggregation}
\label{append:exp-scale-asyn}
\begin{figure}[h!]
	\centering
	\includegraphics[width=0.4\textwidth]{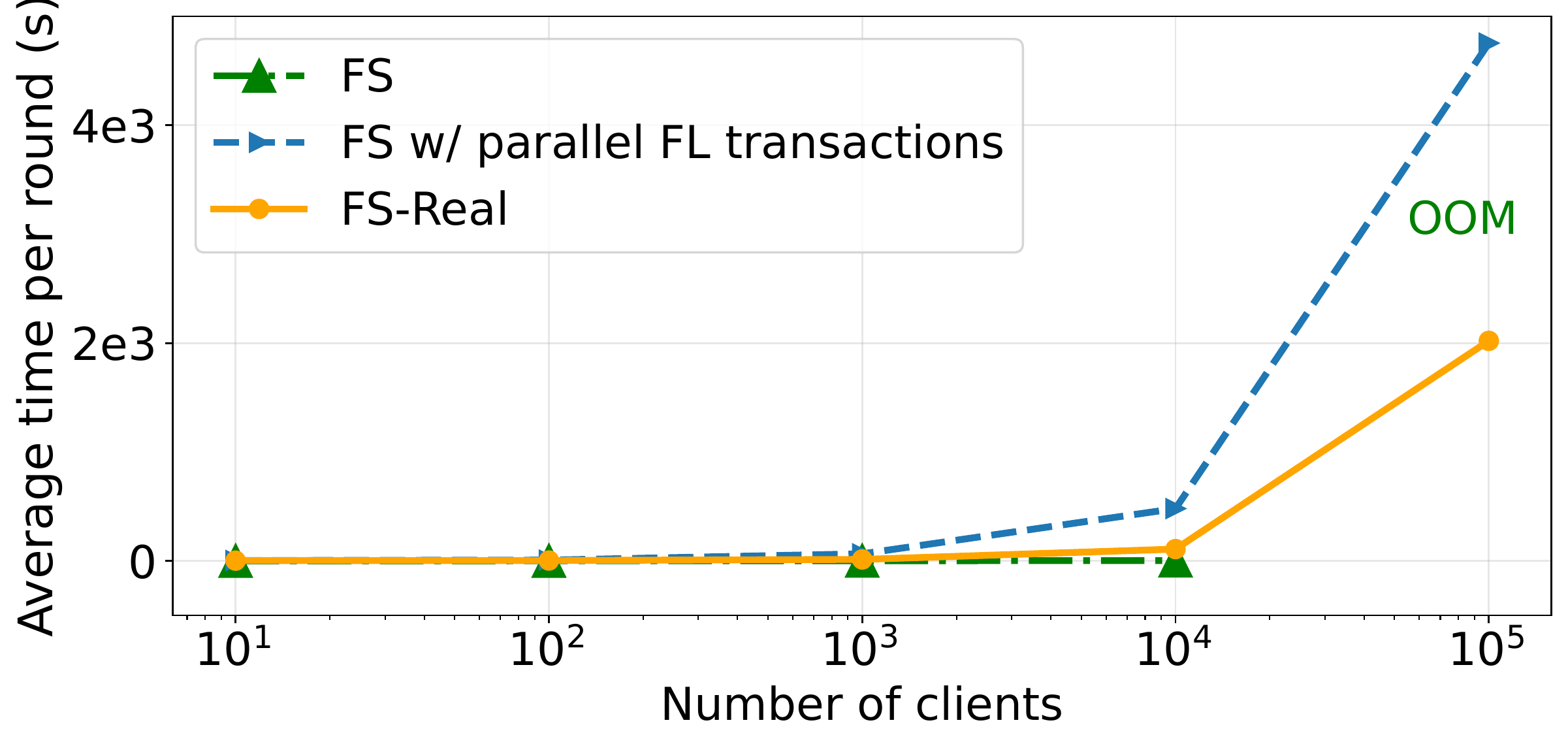}
	\caption{The scalability study of \oursys in asynchronous aggregation FL.} 
	\label{fig:our-scale-asyn}
\end{figure}

In Section \ref{exp:scale}, we have studied the scalability of \oursys that is enhanced by two optimizations, the \textit{parallel FL transactions}, and the \textit{parallel communications}. 
In fact, these two optimizations will bring additional benefits in asynchronous aggregation FL.
We adopt the same experimental settings as Section \ref{exp:scale} to compare the training time at different FL scales, except that the FL algorithm is replaced from synchronous FedAvg to asynchronous FedBuff.
The results are shown in Figure \ref{fig:our-scale-asyn} and we do not compare to FedScale here since it does not natively support asynchronous FL.
Notably, the training time per round of FS with parallel FL transactions is 1.18x $\sim$ 4.77x longer than \oursys, while FS stills failed in the scale of 100,000 due to the OOM.

\begin{table*}[ht]
    \centering
        \caption{Experimental results of personalized FL algorithms on FEMNIST under \textit{Near-Normal} and \textit{Homo-device} distribution.}

    \begin{tabular}{llcccccccccccc}
        \toprule
        \multirow{2}{*}{Distribution}&    \multirow{2}{*}{Method}& \multicolumn{4}{c}{\# Clients = 225} & \multicolumn{4}{c}{\# Clients = 450} & \multicolumn{4}{c}{\# Clients = 900} \\
        \cmidrule(lr){3-6}  \cmidrule(lr){7-10}  \cmidrule(lr){11-14}
        &   & $\overline{acc}$~ & $\protect\widebreve{acc}$ & $\sigma_{acc}$ & $T_{conv}$ & $\overline{acc}$ & $\protect\widebreve{acc}$  & $\sigma_{acc}$ & $T_{conv}$ & $\overline{acc}$ & $\protect\widebreve{acc}$  & $\sigma_{acc}$ & $T_{conv}$ \\ 
        \midrule
        \multirow{3}{*}{\textit{Near-normal}}&  FedAvg & 70.21 &  54.23 &   13.17   & 0.42 & 74.36 &    55.21   &   13.65   & 1.47 & 75.70 &  57.27   &  13.01    & 1.89 \\ 
        &   FT & 72.62 &  54.05 &  16.24    & 0.76 & 75.16 &  58.82  &  15.05  & 1.72 & 75.36 &  59.38 &  15.22    & 1.96 \\ 
        &   FedBABU & 74.03 &  55.39    &  18.43    & 0.97 & 75.01 & 58.57   &  15.95    & 1.16 & 75.83 &  58.29   &  17.15    & 1.48 \\
        \midrule
         \multirow{3}{*}{\textit{Homo-device}}&  FedAvg &   71.35&   54.96&   15.34&   0.54&   73.19   &   55.94   &   13.34   &   1.53    &   75.14   &   59.54   &   12.19   &   1.71  \\
         &  FT &   75.43&   44.50&   17.79&   0.62&   75.28&   47.02&   17.12&   1.76&   76.61&   52.81&  17.00&    1.25  \\
         &  FedBABU &   75.55&   53.12&  17.51&   0.65&   76.13
         &   51.78&   16.86&   1.72&   77.24&   53.33&   15.13&  1.64  \\
         \bottomrule
    \end{tabular}
    \label{tab:homo-hetero-diff-pfl}
\end{table*}

\begin{table*}[ht]
    \centering
        \caption{Experimental results of compression techniques on FEMNIST under \textit{Near-Normal} and \textit{Homo-device} distribution.}

    \begin{tabular}{llccccccccc}
		\toprule                      
	\multirow{2}{*}{Distribution}&	\multirow{2}{*}{Method} & \multicolumn{3}{c}{\# Clients = 225} & \multicolumn{3}{c}{\# Clients = 450} & \multicolumn{3}{c}{\# Clients = 900}  \\
		\cmidrule(lr){3-5}  \cmidrule(lr){6-8}  \cmidrule(lr){9-11}
		& & $\overline{acc}$  & net. traffic & $T_{conv}$  &   $\overline{acc}$     & net. traffic &   $T_{conv}$ &  $\overline{acc}$ &   net. traffic &  $T_{conv}$  \\ \midrule

        \multirow{4}{*}{\textit{Near-normal}}&   FedAvg w/o comp.    &   70.21  &    135.08  &   0.42    &   74.36  &   106.02   &   1.47     &  75.70   &  133.34   &   1.89    \\ 
		& +Gzip &   70.43  &   90.24   &   0.54   &  73.75  &   93.78 &  0.98 &   75.36  &    94.62  &   2.12    \\ 
		& +FP16  &   69.63    &   56.23 &   0.44    &   73.99 &   64.38    &   0.78      &   74.58   & 88.71   &   1.16      \\ 
		& +INT8   &   65.58  &   32.39   &   0.41    &   72.63   &   42.58  &   0.63   &   74.46   &   43.29   &   1.19     \\  
        \midrule
        \multirow{4}{*}{\textit{Homo-device}}&   FedAvg w/o comp.&   71.35&   130.44&   0.54&   73.19&   110.14&   1.53&   75.14& 144.45&   1.71    \\
        &   +Gzip&   70.54&   83.45&   0.64&   72.46&   87.74&   0.72&   74.60&   87.99&  0.74    \\
        &   +FP16&   68.33&   51.95&   0.49&   72.05&   59.22&  0.75&   72.31&  78.84&  0.69    \\
        &   +INT8&   64.50&   30.74&   0.38&   69.43&   40.16&   0.67&   71.45& 50.42&  0.73    \\
        \bottomrule
    \end{tabular}
    \label{tab:homo-hetero-diff-compre}
\end{table*}

\begin{table*}[ht]
    \centering
        \caption{Experimental results of asynchronous aggregation on FEMNIST under \textit{Near-Normal} and \textit{Homo-device} distribution.}

    \begin{tabular}{llcccccccccccc}
		\toprule                       
		\multirow{2}{*}{Distribution}&\multirow{2}{*}{Method} & \multicolumn{4}{c}{\# Clients = 225} & \multicolumn{4}{c}{\# Clients = 450} & \multicolumn{4}{c}{\# Clients = 900}  \\
		\cmidrule(lr){3-6}  \cmidrule(lr){7-10}  \cmidrule(lr){11-14}
		& & $\overline{acc}$ & $\protect\widebreve{acc}$ & $\sigma_{acc}$ & $\overline{uti} \pm  \sigma_{uti}$ &   $\overline{acc}$ & $\protect\widebreve{acc}$ & $\sigma_{acc}$ &       $\overline{uti} \pm  \sigma_{uti}$  &  $\overline{acc}$ & $\protect\widebreve{acc}$ & $\sigma_{acc}$ &       $\overline{uti} \pm  \sigma_{uti}$ \\ \midrule
        \multirow{2}{*}{\textit{Near-normal}}& Sync   &   70.21  &    54.23  & 13.17  &   41.68$\pm$18.87&   74.36  &   55.21  & 13.65  &   27.63$\pm$3.98    &  75.70   &  57.27  & 13.01   &   24.33$\pm$3.72   \\ 
	&	Async       &   80.81  &  67.65  & 10.01   &   54.93$\pm$3.81  &   77.97   & 62.50  & 11.77  &   72.84$\pm$9.14    &   78.34   & 63.01 &  14.57  &   79.41$\pm$4.31    \\ 
        \midrule
        \multirow{2}{*}{\textit{Homo-device}} & Sync &   71.35&   54.13&   15.34&   46.64$\pm$8.34&   73.19&   55.94&  13.34&   25.55$\pm$3.86&   75.14&   59.54&   12.19&    33.89$\pm$5.85   \\
        & Async &   80.17&   63.87&   11.58&   89.23$\pm$7.85&   80.05&   61.55&   12.31&   81.68$\pm$16.59&   78.21&   55.03&   15.22&   44.47$\pm$5.54   \\
        \bottomrule
    \end{tabular}
    \label{tab:homo-hetero-diff-asyn}
\end{table*}

\subsection{Homo-Hetero Gap Study for Advanced FL Features}
\label{append:homo-hetero-advanced-fl}
In Section \ref{sec:exp-gap}, we empirically show that there are substantial performance gaps for FedAvg between homogeneous and heterogeneous device distributions, in terms of various aspects such as accuracy, time to convergence, network traffic, and client utilization. 
In this section, we investigate whether the gap still exists for the advanced FL features within \oursys.
Specifically, we conduct experiments using the same baseline settings we adopted in Section \ref{exp:advanced-fl-tech}, and present the results in Table \ref{tab:homo-hetero-diff-pfl} for personalization, Table \ref{tab:homo-hetero-diff-compre} for communication compression, and Table \ref{tab:homo-hetero-diff-asyn} for asynchronous FL.

From the personalization results (Table \ref{tab:homo-hetero-diff-pfl}), we observe that the runs on \textit{Near-normal} distribution usually gain lower $\overline{acc}$ while higher $\protect\widebreve{acc}$ and $\sigma_{acc}$ than the runs on \textit{Homo-device}. For example, on \textit{Near-normal} distribution, FedBABU achieves 1.12\% $\sim$ 1.52\% lower $\overline{acc}$, while 2.27\% $\sim$ 6.79 \% higher $\protect\widebreve{acc}$ than the ones on \textit{Near-normal} distribution.
This suggests an enhancing effect of device heterogeneity on the bias associated with personalization, which is more severe than that demonstrated in existing homogeneous research work.

While for the compression results (Table \ref{tab:homo-hetero-diff-compre}), in a nutshell, when using the same advanced FL algorithms and parameter settings, the runs on \textit{Near-normal} distribution gain better accuracy than the runs on \textit{Homo-device} in the vast majority of cases.
Taking the INT8 quantization as an example, on \textit{Near-normal} distribution, it gains on average 2.43\% $\overline{acc}$ higher than the ones on \textit{Homo-device} distribution, indicating the great potential of compression technology when device heterogeneity is taken into account.

As for the asynchronous FL (Table \ref{tab:homo-hetero-diff-asyn}), we can see that compared with the runs on \textit{Homo-device}, the runs on \textit{Near-normal} distribution gain lower client utilization at the small device scales ($-43.3$ and $-8.84$ $\overline{uti}$ difference when the number of clients is 225 and 450 respectively), while higher client utilization at the large device scale ($+$ 34.94 $\overline{uti}$ when the number of clients is 900). 
The training acceleration brought by asynchronous FL is more and more obvious on a larger scale under heterogeneous devices, enlightening us on the applicable scenarios and potential improvement space of asynchronous aggregation in real FL applications.

\subsection{End-to-End Evaluation}
\label{append:end2end}
In section \ref{exp:advanced-fl-tech}, we study several advanced FL techniques with \oursys respectively. 
In this section, we examine the end-to-end performance on Twitter dataset by leveraging all the introduced advanced techniques, and summarize the results in Table \ref{tab:end-to-end-res}.
Interestingly, we can find that when comparing the all-in-one method (the last line) with the methods ablating one of the three techniques (the middle three lines), all the three techniques have yielded corresponding gains in the metrics they excel at, \ie, $\overline{acc}$ for personalization, network traffic for INT8 quantization, and client utilization for asynchronous FL.
However, simple quantization still drags down the entire system (compared to the first line), as personalization at this point brings no greater improvement than the model degradation brought by the INT8 technique, which reveals that there is still much room for exploration in efficient personalization techniques.

\begin{table*}[h!]
    \centering
    \caption{The end-to-end evaluation of FS-Device on Twitter dataset under \textit{Near-normal} device distribution.}
    \label{tab:end-to-end-res}
    \begin{tabular}{lccccccccc}
        \toprule                       
	\multirow{2}{*}{Method} & \multicolumn{3}{c}{\# Clients = 250} & \multicolumn{3}{c}{\# Clients = 500} & \multicolumn{3}{c}{\# Clients = 1,000}  \\
	\cmidrule(lr){2-4}  \cmidrule(lr){5-7}  \cmidrule(lr){8-10}
	& $\overline{acc}$ & net. traffic & $\overline{uti} \pm \sigma_{uti}$ &   $\overline{acc}$ &  net. traffic&     $\overline{uti} \pm \sigma_{uti}$  &  $\overline{acc}$ &  net. traffic&    $\overline{uti} \pm \sigma_{uti}$ \\ \midrule
    FedAvg, Syn, w/o INT8 &	61.02&	3.14&95.45  $\pm$  14.87&	62.44&	4.65&	120.84  $\pm$ 18.59&   63.93&  6.09&    123.88 $\pm$ 19.01	\\ \hline
	+FT, +Asyn&	70.26&	2.04&	131.86 $\pm$   12.59&	70.83&	2.82&   159.79 $\pm$ 15.98&    70.74&  4.58&  78.09 $\pm$  12.32  	\\ \hline
	  +FT, +INT8&	53.94&	2.16&	78.09 $\pm$	12.32&    53.93&  2.98&  99.80 $\pm$  15.92&  46.09&	3.59&  111.56 $\pm$   17.10	\\ \hline
	+INT8, +Asyn&   52.45&   1.89& 142.41 $\pm$   14.44&   53.15&   2.52&  137.31 $\pm$    17.55&  53.23   &   3.55&    142.90  $\pm$   19.91	\\ \hline		
	\textbf{+FT, +Asyn, +INT8}&	53.31&	1.91&  134.13 $\pm$    17.78&  53.57&  2.49&	 139.23 $\pm$	16.98& 51.35   &3.58 &  149.13  $\pm$ 19.24	\\    
        \bottomrule
    \end{tabular}
\end{table*}

\end{document}